%% file: ieee_ect_2024.tex
\documentclass[lettersize,journal]{IEEEtran}
\usepackage{amsmath,amsfonts}
\usepackage{algorithmic}
\usepackage{algorithm}
\usepackage{array}
\usepackage[caption=false,font=normalsize,labelfont=sf,textfont=sf]{subfig}
\usepackage{textcomp}
\usepackage{stfloats}
\usepackage{url}
\usepackage{verbatim}
\usepackage{cite}
\usepackage{soul}
\usepackage{color}
\usepackage[dvipsnames, svgnames, x11names]{xcolor}
\usepackage{tikz}
\usepackage{pgfplots}
\usepackage{graphicx}
\usepackage{subfig}
\usepackage{arydshln}
\usetikzlibrary{calc}

\newcommand\LegendImage[1]{
	\draw[%
	/pgfplots/mesh=false,%
	bar width=3pt,%
	bar shift=0pt,%
	mark repeat=2,%
	mark phase=2,#1]
	plot coordinates {
		(0cm,0cm)
		(0.3cm,0cm)
		(0.6cm,0cm)%
	};
}
\newcommand\LegendEntry[1]{\node[anchor=west,black,font=\footnotesize,inner xsep=2pt]{#1};}

\definecolor{mygreen}{RGB}{46,139,87}
\definecolor{myred}{RGB}{255,152,150}
\definecolor{myblue}{RGB}{30,144,255}

\usepackage[colorlinks=true, linkcolor=black, citecolor=black, urlcolor=black]{hyperref}

\usepackage{multirow}
\usepackage{booktabs}
\usepackage{nicematrix}

\usepackage{amssymb}
\usepackage{bm}

\begin{document}

\title{Learning Evaluation Models from Large Language Models for Sequence Generation}

\author{Chenglong Wang, Hang Zhou, Kaiyan Chang, Tongran Liu, Chunliang Zhang, Murun Yang, \\ Quan Du, Tong Xiao, Yue Zhang, \textit{Senior Member, IEEE}, and Jingbo Zhu

\thanks{Chenglong Wang, Hang Zhou, Kaiyan Chang, and Murun Yang are with the Natural Language Processing Laboratory, School of Computer Science and Engineering, Northeastern University, Shenyang 110819, China (e-mail: clwang1119@gmail.com; ctrl.hang@gmail.com; changkaiyan1027@outlo-ok.com; yangmurun@outlook.com). Chenglong Wang and Hang Zhou contributed equally to this work. } 
\thanks{Tongran Liu is with the CAS Key Laboratory of Behavioral Science, Institute of Psychology, Chinese Academy of Sciences, 16 Lincui Road, Chaoyang District, 100101, Beijing, China (e-mail: liutr@psych.ac.cn).}
\thanks{Chunliang Zhang, Quan Du, Tong Xiao (corresponding author) and Jingbo Zhu are with the Natural Language Processing Laboratory, School of Computer Science and Engineering, Northeastern University,
Shenyang 110819, China, and also with the NiuTrans Research, Shenyang 110004, China (e-mail: zhangchunliang@mail.neu.edu.cn;
duquanneu@outlook.com;
xiaotong@mail.neu.edu.cn; zhujingbo@mail.neu.edu.cn).}
\thanks{
Yue Zhang is with the School of Engineering, Westlake University, Hangzhou 310058, China (e-mail: zhangyue@westlake.edu.cn).}
}

\markboth{Journal of IEEE/ACM Transactions on Audio, Speech and Language Processing,~March,~2024}
{Shell \MakeLowercase{\textit{et al.}}: A Sample Article Using IEEEtran.cls for IEEE Journals}


\maketitle

\begin{abstract}
Automatic evaluation of sequence generation, which has traditionally relied on metrics such as BLEU and ROUGE, often struggles to capture the semantic accuracy of generated text due to an overemphasis on n-gram overlap. A promising solution to this issue is the development of model-based metrics, such as BLEURT and COMET. However, these approaches are typically limited by the scarcity of labeled evaluation data, which is essential for training evaluation models. In this work, we address this challenge by proposing the Customized Sequence Evaluation Metric (CSEM), a three-stage model training method that leverages large language models to generate labeled data for metric development, eliminating the need for human-labeled data. Furthermore, we extend the capabilities of CSEM to support a range of evaluation types, including single-aspect, multi-aspect, reference-free, and reference-based evaluations. This flexibility allows for the customization of metrics to fit various real-world scenarios. Experimental results on the SummEval benchmark demonstrate that CSEM can effectively train an evaluation model without human-labeled data. Additional experiments in reinforcement learning and reranking show that metrics developed through CSEM outperform traditional evaluation metrics, leading to significant improvements in sequence quality, as assessed by both commonly used metrics and ChatGPT.  The code is publicly available at \url{https://github.com/wangclnlp/CSEM}.
\end{abstract}


\section{Introduction}
\IEEEPARstart{A}{utomatic} evaluation of sequence generation has traditionally depended on metrics such as BLEU \cite{papineni2002bleu} and ROUGE \cite{lin2004rouge} for tasks including machine translation and summarization. These metrics evaluate generated sequences by comparing them to references based on n-gram or other word-level overlap. However, such overlap-based metrics have well-known limitations in capturing the actual semantic meaning of the output sequence \cite{zhang2019bertscore,wieting2019beyond,sellam2020bleurt}. Notably, they lack robustness against variations in lexical choice or syntax. For example, synonymous words (e.g., \textit{utilize} vs. \textit{use}) are often incorrectly marked as errors due to their failure to match accurately.

To address this limitation, recent efforts have been made to develop model-based evaluation metrics. One common approach is to use pre-trained language models or neural embeddings to train evaluation models that function as metrics, thereby improving the capture of semantic similarities. Examples include BLEURT \cite{sellam2020bleurt}, BARTScore \cite{yuan2021bartscore}, and COMET \cite{rei2022comet}. While these model-based metrics are notably promising and exhibit improved correlation with human judgment compared to traditional metrics such as BLEU and ROUGE, their implementation is not a low-hanging fruit in different real-world scenarios. They typically face the difficulty of training an evaluation model due to the scarcity of human-labeled evaluation data \cite{rei2022comet}. 

Recent successes in large language models (LLMs) have demonstrated that we can annotate data using an LLM, which can then be effectively utilized to train task-specific models without human-labeled data \cite{wang2022self,nijkamp2022codegen,xiao2025foundationslargelanguagemodels}. Motivated by this, we hypothesize that the capabilities of LLMs can be leveraged to create labeled data for training an evaluation model, thereby eliminating the reliance on human-labeled data. As a result, we propose a three-stage training process for the evaluation model and introduce the LLM-based data annotation to develop \textbf{\underline{C}}ustomized \textbf{\underline{S}}equence \textbf{\underline{E}}valuation \textbf{\underline{M}}etric (CSEM). Specifically, the first stage generates diverse evaluation data without labels by sampling from generative models, ensuring data variety for subsequent stages. Secondly, given an LLM with evaluation capabilities, we annotate the generated data to create labeled evaluation data. Finally, we use this labeled data to train the evaluation model with a lightweight language model, ensuring that the resulting model can perform fast and efficient automatic evaluations. Furthermore, we expand the scope of CSEM to include various types of evaluations for sequence generation by modifying both the annotation and model training stages. This expansion allows us to customize evaluation metrics for various real-world scenarios.

In the experiment, we strive to answer the following two key research questions for CSEM. (RQ1): Which method of evaluation data annotation is more effective in CSEM? Commonly used methods for evaluation data annotation typically fall into two categories: \textit{rating} and \textit{comparison ranking}. Through examining the correlation with human judgment, we find that rating is a more effective way of synthesizing evaluation data. (RQ2): Can CSEM be adapted to develop evaluation metrics for different scenarios? We demonstrate that CSEM can effectively develop model-based metrics for a variety of scenarios, including single-aspect, multi-aspect, reference-free, and reference-based evaluations. Furthermore, we conduct experiments to verify the effectiveness of these metrics by using them to provide rewards in RL and reranking. The results show that these metrics significantly improve the performance of generation models and yield more accurate rewards, underscoring the effectiveness of CSEM.

Our contributions are threefold: 
\begin{itemize}
    \item We propose the CSEM method, which effectively creates labeled evaluation data via LLM-based annotation, enabling the development of model-based metrics without the need for human-labeled data.
    \item We expand the scope of CSEM to include various types of evaluations for sequence generation, which enable the customization of a range of sequence metrics based on LLMs. We also thoroughly explore these metrics to provide rewards that improve sequence generation models through RL and reranking.
    \item Through extensive experiments, we demonstrate the effectiveness of CSEM. The results show that CSEM is highly effective. Notably, without relying on human-labeled data, CSEM achieves a 4\% absolute improvement in Spearman correlation while also reducing the number of model parameters by 69\% on the SummEval benchmark, compared to the strong BARTScore \cite{yuan2021bartscore}. Besides, through experiments on RL and reranking, we demonstrate the performance of CSEM in improving sequence models. For example, compared to the baseline, employing our metrics to provide rewards in RL and reranking yields +7.46 COMET points and +3.50 ChatGPT points on the machine translation task.
\end{itemize}

\section{Related Work}
This work builds on related work in data synthesis and automatic evaluation by leveraging LLMs. It is also related to previous work on capability transfer from LLMs.

\textit{Data Annotation via LLMs:}
The data annotation using LLMs has become a critical method for addressing the challenges associated with the scarcity of high-quality, human-labeled data, which is essential for training models in supervised learning contexts \cite{schick2021generating,nijkamp2022codegen,wang2024survey}. Recent studies have primarily highlighted the capability of LLMs to annotate data that replicates not only real-world data distributions but also introduces beneficial variations that significantly enrich model training datasets. These methods are instrumental in extending the capabilities of LLMs across a spectrum of tasks without the traditional reliance on extensive human-labeled data \cite{austin2021program}. However, the application of LLM-based data annotation to develop model-based metrics remains largely unexplored.

\textit{Automatic Evaluation via LLMs:}
Recent works rely on the in-context learning capabilities of the LLM and prior knowledge of human behaviour to investigate its application to the evaluation of sequence generation tasks. This core idea is to directly use an LLM as an evaluator by feeding the task description and the evaluation criteria as a prompt. These works could be classified into two groups. The first group focused on verifying the effectiveness of the LLM-based evaluation for various sequence generation tasks \cite{fu2023gptscore, kocmi2023large, wang2023chatgpt, lai2023multidimensional}. For example, \ \cite{kocmi2023large} demonstrated the effectiveness of applying LLMs to translation evaluation by computing the correlation scores between LLMs and humans. The second group tended to improve the LLM evaluation capability, such as incorporating chain-of-thoughts \cite{liu2023gpteval, luo2023chatgpt} and roleplayers \cite{wu2023large,chan2023chateval}. However, despite the powerful evaluation capabilities of LLMs, directly using them for evaluation can be computationally expensive. Unlike these approaches, our work investigates the use of LLM-based evaluation capabilities to annotate data for training evaluation models, addressing the scarcity of human-labeled data. As another bonus, the resulting evaluation model offers significantly lower operational costs and faster evaluation speeds.

\textit{Transferring Capability from LLMs:}
A commonly used technique to transfer capabilities from LLMs is knowledge distillation \cite{lin2020weight,tripathi2023divide,wang2023improved}. Its core idea is to transfer the knowledge from an LLM into a lightweight model. For example, \cite{ho2022large} transferred the reasoning capability from PaLM \cite{chowdhery2022palm} to T5 \cite{raffel2020exploring} by using sequence-level distillation \cite{kim2016sequence}. Furthermore, when transferring the task-specific capability from an LLM, \cite{hsieh2023distilling} proposed a step-by-step distillation mechanism, which learns from both LLM-labeled labels and rationales based on multi-task learning. In this work, we focus on transferring the evaluation-related capacity of an LLM for large-scale applications, a previously unexplored research topic. The few related ones mainly focused on learning preference models from LLMs \cite{lee2023rlaif, cui2023ultrafeedback}. Specifically, these works used the trained generation models to generate multiple sequences for each input and simulated human preference data (i.e., comparisons of the sequences) with an LLM. Based on this preference data, the preference models can be learned via ranking loss \cite{ouyang2022training}. However, this process only transfers the sequence comparison capability of an LLM, but not its sequence evaluation capability (see Section \ref{sec:comparison-rating-ranking}) \cite{ziya-reward-7B}.

To our knowledge, we are the first to conduct such a comprehensive exploration of using LLMs to develop a model-based metric without relying on human-labeled data. The most closely related works primarily focus on utilizing LLMs to generate synthetic datasets of sentence pairs with similarity ratings for translation quality evaluation \cite{mohtashami2023learning}. Another related one involves training evaluation models using unsupervised training \cite{thompson2020automatic,fomicheva2020unsupervised}. In contrast to these works, we directly use LLMs to annotate diverse labeled data for the development of model-based metrics across various generative tasks.

\vspace{-2mm}
\section{Preliminaries}
\subsection{Sequence Generation Model}
Given an input $x$ such as a text, a sequence generation model generates a sequence of $N$ tokens $y = \{y_{1},\dots, y_{N}\}$, where each token $y_{t}$ is drawn from a predefined vocabulary.
At the training stage, the model learns the probability
\begin{eqnarray}
		\mathrm{Pr}_{\theta}(y|x) &=& \prod_{t=1}^{N}\mathrm{Pr}_{\theta}(y_{t}|y_{<t},x)
\label{eq-generation-pro}
\end{eqnarray}
where $y_{<t}$ is the prefix $\left\lbrace y_{1}, y_{2}, \dots, y_{t-1}\right\rbrace $, and $\theta$ is a set of model parameters.
In this process, the standard training objective is to maximize the likelihood over all the tokens of the target sequence, \textit{i.e., maximum likelihood estimation (MLE)} \cite{myung2003tutorial}.
At the inference stage, we generate tokens sequentially according to the probability $\mathrm{Pr}_{\theta}$.

\subsection{Training Evaluation Models}
\label{sec:background-training-evaltaion-models}
In the evaluation of sequence generation tasks, while traditional metrics such as BLEU and ROUGE metrics are commonly utilized, there is also a significant opportunity to train an evaluation model that functions the metric that can offer a deeper and potentially more accurate measure of the quality of generated sequences. This approach involves using pre-trained language models as encoders to capture the nuances in both the input $x$ and the generated sequence $\hat{y}$, subsequently mapping these representations to a numerical score $\mathcal{M}(x,\hat{y})$ that serves as the quality score for sequence $\hat{y}$, where $\mathcal{M}(\cdot)$ denotes the evaluation model. Additionally, in cases where a reference sequence $y_{\mathrm{ref}}$ exists, it can also be encoded simultaneously to train a reference-based evaluation model $\mathcal{M}(x,\hat{y}, y_{\mathrm{ref}})$. To train the evaluation model, we can collect labeled evaluation data on a set of generated sequences in two primary formats: \textit{rating} and \textit{comparison ranking}. The first is often a continuous or discrete numerical value, such as a score on a scale (e.g., 1-5 stars or 1-10 points). In this case, we can give the loss function through the mean squared error:
\begin{eqnarray}
     \mathcal{L}_{r} &=& - \mathbb{E}_{(x,\hat{y},l_r)\sim \mathcal{D}_{r}} \big(\mathcal{M}(x,\hat{y})- l_r\big)^2
\end{eqnarray}
where $\mathcal{D}_{r}$ denotes the dataset of training the evaluation model and $l_r$ denotes the labeled rating for $\hat{y}$. The second is a ranking. In this approach, two different sequences $\hat{y}_a$ and $\hat{y}_b$ are generated, and the labeler ranks them, for example, $\hat{y}_a \succ \hat{y}_b$ indicating that $\hat{y}_a$ is better than $\hat{y}_b$. The loss function can given through a pair-wise ranking:
\begin{eqnarray}
    \mathcal{L}_{c} &=& - \mathbb{E}_{(x,\hat{y}_{a},\hat{y}_{b})\sim \mathcal{D}_{c}} \log \sigma(\mathcal{M}(x,\hat{y}_{a}) - \mathcal{M}(x,\hat{y}_{b}))
    \label{eq:ranking-loss}
\end{eqnarray}
where $\sigma$ denotes the sigmoid activate function and $\mathcal{D}_{c}$ denotes a set of comparison ranking data. This training approach is also the current method used in reward modeling when aligning LLMs with human preferences \cite{ouyang2022training}.

\subsection{Reinforcement Learning}
The optimization objective of RL for sequence generation models is to maximize the long-term reward, written as $\arg\max_{\theta}\mathbb{E} [r(\hat{y})]$, where $r(\cdot)$ is a reward function computing long-term reward for $\hat{y}$.
$r(\cdot)$ is typically defined by an evaluation metric.
To achieve this objective, policy gradient methods, such as REINFORCE \cite{williams1992simple} and minimum risk training (MRT) \cite{shen2015minimum}, are often used. 
Specifically,  REINFORCE uses log derivatives to define the loss function:
\begin{eqnarray}
		\mathcal{L}_{\mathrm{REINFORCE}} &=&  -\mathbb{E}_{\hat{y} \sim \Omega(x)}\log \mathrm{Pr}_{\theta}(\hat{y}|x) r(\hat{y})
 \label{eq-pg}
\end{eqnarray}
where $\Omega(x)$ is an approximated space of sampling, and it consists of the sampled sequences.
Furthermore, MRT uses these sampled sequences to approximate a posterior distribution with renormalization and gives a new loss function:
\begin{eqnarray}
		\mathcal{L}_{\mathrm{MRT}} &=& \mathbb{E}_{\hat{y}\sim \Omega(x)}Q_{\theta}(\hat{y}|x)\big[-r(\hat{y})\big]
	\label{eq-mrt}
\end{eqnarray}
where $Q_{\theta}(\hat{y}|x)$ is a distribution defined on the approximated space, and it can be defined by
\begin{eqnarray}
		Q_{\theta}(\hat{y}|x) &=& \frac{\mathrm{Pr}_{\theta}(\hat{y}|x)^{\alpha}}{\sum_{\hat{y}\in \Omega(x)}\mathrm{Pr}_{\theta}(\hat{y}|x)^{\alpha}}
\end{eqnarray}
where $\alpha$ is a smoothness parameter.
Based on the posterior distribution, MRT can achieve better performance compared with REINFORCE \cite{kiegeland2021revisiting, donato2022mad}.

\subsection{Reranking}
Reranking refers to reordering or reevaluating a set of candidate sequences generated by a trained model \cite{shen2014dependency, lee2021discriminative, liu2021addressing}.
Given a set $\mathcal{Y}$ of $N$ candidate sequences for input, we also use the reranking approach to maximize the reward computed without the corresponding references.
Typically, we can use the reward model to score the candidate 
sequences and pick a final generated sequence that has a maximum reward score:
\begin{eqnarray}
    \hat{y}_{\mathrm{RR}} &=& \arg\max_{\hat{y}\in\mathcal{Y}}r(\hat{y})
\end{eqnarray}
When multiple rewards $(r_{1}, r_{2}, \cdots, r_{J})$ are used, we can pick a sequence through assigned weights $(w_{1}, w_{2}, \cdots, w_{J})$, where $J$ is the number of rewards:
\begin{eqnarray}
		\hat{y}_{\mathrm{RR}} &=& \arg\max_{\hat{y}\in\mathcal{Y}} \sum_{j=1}^{J} w_{j} \times r_{j}(\hat{y})
\end{eqnarray}

Note that the use of appropriate evaluation metrics to provide rewards is crucial for the success of applying RL and reranking to improve sequence generation models \cite{shu2021reward, fernandes2022quality}. However, this also poses significant challenges in real-world scenarios. Notably, we can not always straightforwardly write a function that can evaluate a generated sequence for different sequence generation tasks while simultaneously considering various aspects. While several methods have been proposed for learning evaluation models, they typically require a substantial amount of human-labeled data \cite{zhang2016learning, rei2020comet, fernandes2022quality}.

\section{Evaluation Capability Transfer}
\subsection{Overview}
\begin{figure}[t!]
    \centering
    \includegraphics[scale=0.40]{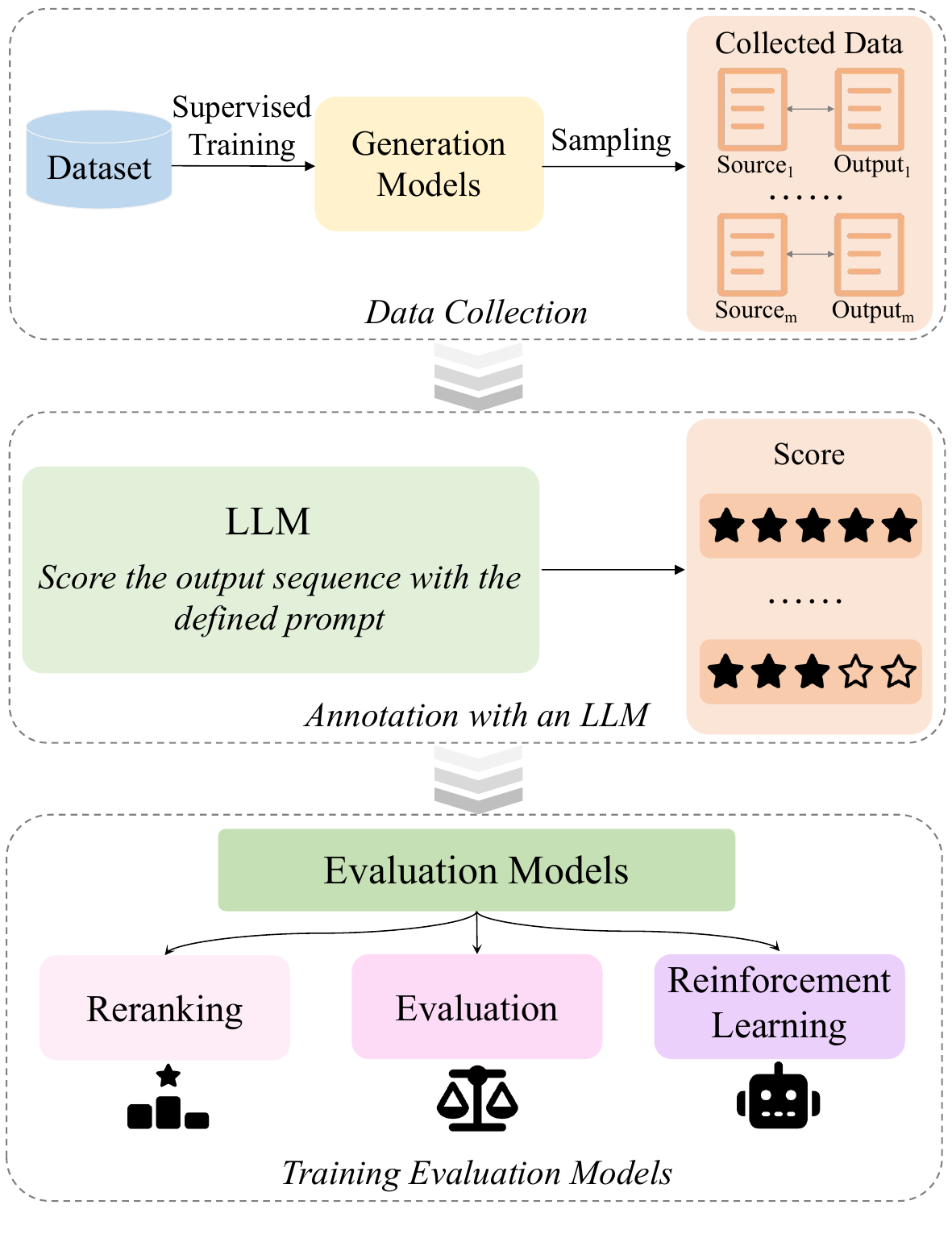}
    \vspace{-3mm}
    \caption{
    An overview of the proposed CSEM. 
    The CSEM consists of three stages: 1) collecting data by sampling from the trained sequence generation model; 2) annotating the evaluation scores for each sample by leveraging an LLM, and 3) training evaluation models with a pre-trained language model.
    }
    \vspace{-4mm}
    \label{fig:main_image}
\end{figure}

In this work, we aim to harness the capabilities of LLMs to annotate labeled data, thereby eliminating the reliance on human-labeled data for the development of model-based metrics. To achieve this, we introduce the CSEM method to accomplish this goal. The structure of CSEM is illustrated in Figure \ref{fig:main_image}. As depicted, CSEM encompasses three primary stages: data collection, annotation with an LLM, and training evaluation models. Additionally, we utilize these metrics to improve sequence generation models. The subsequent subsections provide a detailed explanation of each stage.

\input{use_images/prompt}
\subsection{Data Collection}
\label{sec:data-collection}
To create labeled evaluation data, we initiate the process by training a sequence generation model using MLE on the provided training set. Note that in CSEM, we opt to annotate data using a rating-based format. Consequently, we proceed to generate a set of data pairs $\mathcal{D}_{r}=\{(x^1, \hat{y}^1), \cdots, (x^n, \hat{y}^n)\}$, where $n$ denotes the number of data pairs. This is achieved by randomly selecting $n$ training samples from the training set. The trained generation model is then used to generate a candidate sequence for each selected input. 

Our goal is to produce a diverse array of candidate sequences, enhancing the robustness and generalization ability of the evaluation models to be trained. To this end, we have devised a detailed process to derive the candidate sequence for each input, aiming to optimize the quality and variability of the generated data. Specifically, given an input $x^{i}$, we employ top-$k$ sampling, and sample $j$ output sequences $\{\hat{y}_1^i, \cdots, \hat{y}_j^i\}$ from different checkpoints. Furthermore, the reference $y^{i}_{\mathrm{ref}}$ of $x_{i}$ in the training set is added to the corresponding output sequence set. Finally, we randomly select one sequence $\hat{y}^i$ from $\{\hat{y}_1^i, \cdots, \hat{y}_j^i, y^{i}_{\mathrm{ref}}\}$ as the candidate sequence of $x^{i}$ to form an data pair $(x^{i}, \hat{y}^i)$.

In selecting the data for this stage, we need to ensure that the training of the evaluation model is aligned with the target evaluation scenarios. To achieve this, we carefully choose inputs that match the distribution of the desired evaluation task. For example, in a summarization task, we select inputs from the CNN/DM training set \cite{hermann2015teaching}, as it contains a wide range of generalizable scenes commonly seen in the domain. Additionally, this dataset is frequently used in training sequence generation models, making it highly relevant. However, note that our CSEM method is general and can be applied to any sequence generation task. Therefore, the choice of data can be adjusted arbitrarily based on the specific evaluation scenario, enabling the generation of high-performance evaluation models.

\subsection{Annotation with an LLM}
\label{sec:annotation}
Considering the proven robust evaluation capabilities of LLMs across various generative tasks \cite{fu2023gptscore, kocmi2023large, mohtashami2023learning, wang2023chatgpt, lai2023multidimensional}, we use an LLM to annotate the evaluation score for each data pair. This process is performed after data collection and involves using a natural language prompt that includes the task description and evaluation criteria. Following the methodology in \cite{wang2023chatgpt}, we design all prompts to facilitate a rating scale from one to five stars, as illustrated in Figure \ref{fig:prompt}. To adapt our method to different scenarios, we have expanded it by designing prompts that cater to different types of sequence generation evaluations tailored to suit a range of application scenarios. These evaluation types include single-aspect, multi-aspect, reference-free, and reference-based evaluations. This allows us to select the most appropriate evaluation type based on the specific requirements of the application scenario, thereby enhancing the applicability and effectiveness of CSEM.

There are two key advantages to employing an LLM as an annotator: 1) we can intuitively specify the desired evaluation criteria through explicit natural language description in the prompt, and  2) we can obtain high-quality annotations based on the evaluation and in-context learning capabilities of LLMs, which can be significantly cheaper and faster compared to employing human annotators \cite{ding2022gpt, kang2023distill, chiang2023large}.

Two questions arise here: \textit{why not use a few-shot evaluation template} and \textit{why not use multiple LLMs to annotate evaluation scores?} Regarding the first question, we opted for a zero-shot template as prior studies have demonstrated its effectiveness in sequence evaluation \cite{wang2023chatgpt, liu2023gpteval}. Furthermore, the use of few-shot templates is infrequent in LLM-based evaluations, primarily to avoid the transmission of biases from human evaluators in the demonstrations. Concerning the second question, we experimented with learning evaluation from both ChatGPT and \texttt{text-davinci-003}. However, this approach proved less effective, yielding a Spearman correlation of 0.397 on the SummEval benchmark, which is lower than the 0.442 achieved using only ChatGPT. We attribute this to potential conflicts in the metric scales used by these two models. Consequently, in this work, we have decided against using multiple LLMs to annotate evaluation scores.

\subsection{Training Evaluation Models}
\label{sec:training-eval-models}
Although LLMs have demonstrated strong evaluation capabilities, they are not directly suitable to function as evaluation models for developing model-based metrics. This limitation stems from the requirement for metrics to be reproducible, lightweight, and fast to meet the frequent testing of model-generated quality, especially in scenarios where the test sets may consist of thousands of samples. In this section, we utilize the annotated data to fine-tune relatively lightweight language models (LMs) that serve as evaluation models. The models we ultimately train have parameter sizes of 125M and 355M, which are consistent with the parameter sizes of currently available open-source model-based metrics, such as BERTScore and BARTScore. These parameter sizes allow for a fast run on one GPU with 12G or 8G memory (see Section \ref{sec:reward_query_efficiency}), meeting the speed requirements necessary for large-scale automatic evaluations. Subsequently, we will detail the model architecture and the optimization techniques used in the training of these evaluation models.

\subsubsection{Model Architecture}
\label{sec:model-arch}
Our evaluation model architecture follows COMET \cite{rei2020comet}, which employs the LM as an encoder and the feed-forward network as a regressor. Specifically, we first encode each input sequence (i.e., source, generated sequence, and human reference) using the LM. From the output vectors, we extract the \texttt{[CLS]} embedding, which serves as the representation of the entire sequence. This is based on the consideration that the \texttt{[CLS]} embedding typically captures the rich semantics of the textual output, making it particularly suitable for our evaluation purposes. Following \cite{shimanaka2018ruse}, we define the input of the regressor based on the extracted embeddings:
\begin{eqnarray}
		\begin{aligned}
		\bm{x}_{\mathrm{regressor}}=&[\bm{h}_{\hat{y}};\bm{h}_{y_{\mathrm{ref}}};\bm{h}_{\hat{y}}\odot\bm{h}_{x};\bm{h}_{\hat{y}}\odot\bm{h}_{y_{\mathrm{ref}}};\\&|\bm{h}_{\hat{y}}-\bm{h}_{x}|;|\bm{h}_{\hat{y}}-\bm{h}_{y_{\mathrm{ref}}}|]
		\end{aligned}
\end{eqnarray}
where $\bm{h}_{x}$, $\bm{h}_{\hat{y}}$ and $\bm{h}_{y_{\mathrm{ref}}}$ denote the \texttt{[CLS]} embeddings of source, generated sequence and human reference, respectively. 
When the human reference is absent, i.e., reference-free evaluation, the input of the regressor is
\begin{eqnarray}
		\bm{x}_{\mathrm{regressor}}=[\bm{h}_{\hat{y}};\bm{h}_{x};\bm{h}_{\hat{y}}\odot\bm{h}_{x};|\bm{h}_{\hat{y}}-\bm{h}_{x}|]
\end{eqnarray}
The regressor takes in the given input and outputs a continuous value that serves as a score of the generated sequence. 

In practice, we operate under the assumption that not all information is necessary for the purpose of evaluation. For instance, the source input may be considered superfluous when assessing the fluency of a generated sequence. Including such information could inadvertently introduce noise into the evaluation model. Based on this premise, we streamline the fluency evaluation models by omitting the source input. Specifically, the inputs to the regressor are the embeddings of the generated sequence, $\bm{h}_{\hat{y}}$, and the reference sequence, $\bm{h}_{y_{\mathrm{ref}}}$, which are used to evaluate the quality without the influence of extraneous data.

\subsubsection{Optimization}
We train all evaluation models using a regression loss function that minimizes the mean squared error between the predicted value (i.e., a continuous score) and the annotated score:
\begin{eqnarray}
    \mathcal{L}_{r} &=& - \mathbb{E}_{(x,\hat{y})\sim \mathcal{D}_{r}}\big(\mathcal{M}(x, \hat{y})- S(x, \hat{y})\big)^2
\end{eqnarray}
where $S(\cdot)$ denotes the evaluation score annotated by an LLM. Here, the annotated score, denoted by stars, is converted into a real value through min-max normalization. Additionally, we investigated the use of a classification loss function for training evaluation models, treating each star rating as a distinct class (refer to Section \ref{sec:classification}). The results indicate that a regression loss yields superior performance, as it allows the model to find subtle differences between output sequences.

\input{tables/single_aspect}

\subsection{Applying Evaluation Models}
\label{sec:apply}
We train evaluation models with a small number of parameters, making CSEM metrics highly scalable and well-suited for large-scale applications, such as providing rewards in RL and reranking \cite{he2024improving}.

\subsubsection{Reinforcement Learning}
We employ policy gradient-based RL to optimize for our evaluation model-based metrics.
Following \cite{yehudai2022reinforcement}, we fine-tune the pre-trained sequence generation model with a weighted average of the RL loss $\mathcal{L}_{\mathrm{RL}}$ and the MLE loss $\mathcal{L}_{\mathrm{MLE}}$,  written as
\begin{eqnarray}
    \mathcal{L}_{\mathrm{weighted}} &=& \lambda\times\mathcal{L}_{\mathrm{RL}}+(1-\lambda)\times\mathcal{L}_{\mathrm{MLE}}
    \label{eq:rl-loss}
\end{eqnarray}
where $\lambda$ is a balance factor that is tuned on the validation set.
Here, we compute $\mathcal{L}_{\mathrm{RL}}$ using MRT with Eq. \ref{eq-mrt}.

\subsubsection{Reranking}
The reference-free metrics enable the reranking of candidate sequences for sequence generation models. This enables the selection of the optimal generated sequence to maximize specific evaluation criteria in the prompt used by CSEM. Inspired by \cite{fernandes2022quality}, we use multiple metrics in the reranking process to obtain better sequences. These metrics are developed by CSEM with single-aspect evaluation and focus on different evaluation aspects. To assign a reasonable weight to each metric, we tune weights for these models to maximize a given reference-based metric on a validation set by using Travatar's MERT tool \cite{neubig2013travatar}.

\section{Experiments}
In this section, we verify CSEM in two ways. First, we test the performance of evaluation models trained by CSEM on the SummEval benchmark, thereby confirming its effectiveness in training evaluation models.
Second, we utilize the obtained evaluation model-based metrics to provide rewards for the sequence generation tasks, including machine translation, text style transfer, and summarization. This step verifies whether these metrics can effectively improve sequence generation models through RL and reranking.

\input{tables/multi_aspect}

\input{tables/mt_summary_dataset}

\input{tables/style_transfer_dataset}

\subsection{Implementation Details}
\subsubsection{Datasets}
The datasets used for each sequence generation task were as follows:
\begin{itemize}
    \item \textit{Machine Translation}: 
    We conducted experiments on two machine translation datasets, including a small-scale IWSLT’14 German-English (De-En) dataset and a large-scale WMT’14 English-German (En-De) dataset. We preprocessed the dataset following \cite{hu2021ranknas}.
    \item \textit{Text Style Transfer}:
    The experiments regarding the text style transfer task were conducted on Grammarly’s Yahoo Answers Formality Corpus (GYAFC) dataset \cite{rao2018dear}.
    This dataset is a formality style transfer with parallel informal and formal sentences from two domains: Entertainment \& Music (E\&M) and Family \& Relationships (F\&R).
    \item \textit{Summarization}:
    We also tested the capability of the CSEM to train evaluation models for the summarization task on the CNN/DM dataset \cite{hermann2015teaching}.
\end{itemize}
The statistical information on the utilized datasets is summarized in Tables \ref{tab_statistical_mt_summay_datasets} and \ref{tab_statistical_style_tranfer_datasets}.

\subsubsection{Training Evaluation Models}
We chose ChatGPT (the \texttt{gpt-3.5-turbo-0613} version) as our LLM.
It was fine-tuned by RL from human feedback \cite{christiano2017deep} and proven to have remarkable evaluation capabilities for different sequence generation tasks \cite{wang2023chatgpt}.
For evaluation aspects, we designed one multi-aspect evaluation and four single-aspect evaluations tailored to each task.
The prompts with these designed evaluation aspects are described in Tables \ref{tab:single-aspect} and \ref{tab:multi-aspect}.
For data collection, we randomly selected 15K training samples from the training set and sampled five output sequences for each input. 
For evaluation model architecture, we used RoBERTa-base and RoBERTa-large \cite{liu2019roberta} as the encoder model, respectively.
Note that when dealing with the machine translation task, we used XLM-RoBERTa \cite{conneau2019unsupervised}, a multilingual LM, as the encoder model.
We trained the evaluation model based on the COMET codebase.
The learning rate, the maximum number of epochs, and the batch size were set to 1e-5, 10, and 32, respectively.
We trained all evaluation models with 16-bit floating point precision on one TITAN V GPU.

\subsubsection{Training Sequence Generation Models}
For machine translation and summarization tasks, we pre-trained a standard Transformer base model \cite{vaswani2017attention} using MLE until convergence.
For the text style transfer task, following \cite{lai2021thank}, we fine-tuned a BART base model \cite{lewis2019bart} to serve as our pre-trained generation model.
All models were trained on four TITAN V GPUs.
For RL training, we used the MLE checkpoint with the lowest validation set loss to initialize a sequence generation model.
We sampled five candidate sequences for each source input during RL training.
The balance factor $\lambda$ in Eq. \ref{eq:rl-loss} was 0.7.
For the IWSLT'14 De-En and WMT'14 En-De datasets, we trained the pre-trained models over 10 epochs and 8000 steps, respectively, using a batch size of 4096 tokens (at the token level). Subsequently, we conducted tests on these models using their corresponding datasets; specifically, the IWSLT’14 De-En test set was used to evaluate the IWSLT model, and the WMT’14 En-De test set was employed to evaluate the WMT model.
For the text style transfer and summarization tasks, we trained the pre-trained models for approximately 4000 steps using a batch size of 4096 tokens. Following the methodology outlined in \cite{wang2023esrl}, we adopted a two-stage sampling approach to enhance the efficiency of RL training. Specifically, stage one is to sample the candidate sequences with an autoregressive mode \cite{xiao2023introduction}. Stage two is to compute the probabilities of the sampled candidate sequences and their gradient information. This two-stage sampling approach can take full advantage of the Transformer’s parallelism computation, so the excessive computational graph storage requirements disappear.
For reranking purposes, we generated 50 candidate sequences for each task using the top-$k$ sampling method. During inference, we employed the beam search strategy, adhering to the same hyperparameters as described in \cite{li2022ode}.

\input{tables/summary_multi_aspect}

\input{tables/mt_res}

\subsection{Results of SummEval Benchmark}
We computed the correlation scores of four single-aspect summarization evaluation models with human judgments on the SummEval benchmark \cite{fabbri2021summeval}, respectively. The results are shown in Table \ref{tab:summary-multi-aspect}. We compare CSEM with different evaluation methods that achieved state-of-the-art performance on summarization evaluation, including ROUGE \cite{lin2004rouge}, BERTScore \cite{zhang2019bertscore}, MoverScore \cite{zhao2019moverscore}, and BARTScore \cite{yuan2021bartscore}. The results show that CSEM can achieve optimal correlation results for the average performance compared to all baseline evaluation methods except ChatGPT. Notably, although the evaluation data is labeled by ChatGPT, we also see that CSEM can surpass ChatGPT on some evaluation aspects. For example, the reference-based CSEM-large metric achieves a 0.529 Spearman correlation for the relevance evaluation aspect, which exceeds the ChatGPT (i.e., 0.512). A similar phenomenon can also be found in knowledge distillation \cite{wang2021niutrans,burns2023weak}. We here hypothesize that this improved performance stems from two primary factors. First, although we use LLMs for rating-based annotation, our training model employs a regression approach. This potentially allows the evaluation model to generalize better, capturing subtle differences between sequences that may not be fully captured by LLMs alone. We delve into this in more detail in Section \ref{sec:classification}. Second, the used model architecture based on COMET is specifically designed for evaluation tasks. Unlike the general-purpose LLM architecture, the design of COMET is tailored to capture evaluation-specific features, making it more adept at capturing features relevant to evaluation. Coupled with LLM supervision, this targeted design likely enables weak-to-strong generalization.

Further analysis of the CSEM-base and CSEM-large metrics, particularly for reference-based evaluation, reveals that CSEM-large excels across various correlation metrics. From the results, we can observe that CSEM-large consistently outperforms all baselines except in fluency, where BARTScore-FT surpasses it. However, this can be attributed to the extensive training of BARTScore-FT on the CNN/DM dataset. Interestingly, in reference-free evaluations, CSEM-large does not consistently outperform CSEM-base, suggesting that large model parameters may not be necessary for CSEM in these contexts. Indeed, in some cases, larger parameters could lead to overfitting and, thus, poorer generalization capabilities, as noted in \cite{rei2023scaling}. The results suggest that \textit{for reference-based evaluation scenarios, developing a CSEM-large version yields superior performance. Conversely, for reference-free evaluation scenarios, a CSEM-base version is advisable to achieve a more lightweight model-based metric.}

\input{tables/style_trans_res}

\input{tables/summary_res}

\subsection{Results of Reinforcement Learning and Reranking}
We conducted experiments to further validate the effectiveness of CSEM by using it to provide rewards in RL and reranking. The experiments focus on two main questions: 1) \textit{Can CSEM-based metrics effectively improve sequence generation models?} and 2) \textit{Does CSEM outperform traditional metrics in RL and reranking?} To address these questions, we applied different versions of CSEM in RL and reranking across various model tasks and compared the results with both automatic and human evaluations (Paragraphs 1, 2, 3, and 4). Additionally, we compare the performance of rewards derived from CSEM with those derived from traditional metrics (Paragraph 5).
\subsubsection{Machine Translation}
Table \ref{tab:mt-res} compares the performance of different versions of CSEM metrics when applied to machine translation models. First, compared to MLE, CSEM-based metrics, when applied with RL and reranking, significantly enhance ChatGPT-based metrics. This also serves as evidence that the evaluation models trained by CSEM are well-aligned with ChatGPT. Second, these improvements are consistently observed across other evaluation metrics. Notably, with a single multi-aspect evaluation model, the RL+RR approach achieves an impressive +7.46 COMET-20 point gain on the IWSLT task. One possible explanation for this is that CSEM offers a more comprehensive evaluation, capturing aspects that overlap with other existing metrics. Third, we analyze the performance gains from RL and reranking. The results indicate that reranking provides a more stable optimization process than RL when using a reference-free metric. Additionally, combining both techniques (i.e., RL+RR) further enhances reward optimization, demonstrating the effectiveness of their joint application.

\subsubsection{Text Style Transfer and Summarization}
The results of the text style transfer task are listed in Table \ref{tab:style-transfer-res}. The results further highlight that optimization using our metrics significantly enhances the performance of the text style transfer task. As shown by the comparison between the ``Chat.'' and ``ChatRef.'' columns, a similar trend observed in the machine translation task is also evident here: optimization with our metrics yields a substantial performance boost over MLE. Furthermore, we conduct experiments on the summarization task, as shown in Table \ref{tab:summary-res}. The performance is evaluated using ROUGE-1, ROUGE-2, ROUGE-L, Chat., and ChatRef. metrics. The results also show that optimizing with our metrics yields superior results over MLE.

\input{tables/case_study_IWSLT_WMT}

\subsubsection{Case Study}
Tables \ref{tab:case-study-iwslt-wmt} showcase various case studies from the IWSLT and WMT translation tasks. We selected the MLE and RL+RR with the CSEM-large metric for comparison, owing to its demonstrated superiority in machine translation tasks. The findings indicate that optimizing translations with our metrics significantly outperforms MLE. Notably, our metrics concentrate on aspects of meaning and grammar (as outlined in Table \ref{tab:multi-aspect}), effectively addressing issues related to meaning errors, incompleteness, and grammar errors.

\subsubsection{Human Evaluation} \label{sec:human_evaluation}
To further evaluate the effectiveness of our metrics, we conduct a human evaluation. We employ the metrics developed by CSEM to assign rewards in RL and reranking, aiming to enhance sequence generation models. The performance of these models is then compared to the MLE baseline. For this evaluation, we randomly select 200 source sentences from the IWSLT'14 test set, along with their corresponding translations generated by various models. These sentences are selected with the only constraint that their token length falls between 3 and 25, as in \cite{fernandes2022quality}. Annotators are asked to assign a score from 1 to 5 for each translation based on the following criteria.
\begin{itemize}
    \item \textit{Score 1}: The translation conveys a meaning that is completely different from the source sentence.
    \item \textit{Score 2}: The translation omits significant portions of the source meaning and lacks fluency.
    \item \textit{Score 3}: The translation omits significant portions of the source meaning but maintains fluency, though it contains some grammatical errors.
    \item \textit{Score 4}: The translation conveys the full meaning of the source sentence and reads fluently, though it contains some grammatical errors.
    \item \textit{Score 5}: The translation accurately conveys the meaning of the source sentence, reads fluently, and is free of grammatical errors.
\end{itemize}

\input{tables/human_eval_res}

Table \ref{tab_human_eval} shows the results of the human evaluation for the machine translation task. From these results, we see that optimization using our metrics outperforms the MLE baseline. We also observe that RR, which incorporates multiple evaluation metrics, yields higher translation quality compared to RL, which is combined with RL+RR using only a single evaluation metric. This finding contrasts with the results in Table \ref{tab:mt-res}. One possible explanation for this discrepancy is that RL+RR with a single evaluation metric is more susceptible to overfitting, as highlighted in \cite{fernandes2022quality}. This suggests that the sequence generation model may achieve better performance when trained with RL+RR and multiple evaluation metrics. However, implementing this comes with significant computational costs. We plan to explore ways to balance efficiency and effectiveness in future work.

\subsubsection{Performance on Different Rewards}
To further verify the effectiveness of CSEM, we compare the performance of CSEM with traditional metrics in RL. Specifically, we evaluate the impact of different reference-based metrics, including BLEU, COMET-20, COMET-22, and ChatGPT, on training translation models using RL. The results, as shown in Table \ref{tab:diff_rewards}, show that CSEM outperforms all traditional rewards in human evaluation, except for ChatGPT. Additionally, RL training with CSEM demonstrates strong performance with the ChatRef. metric, indicating a high level of consistency between CSEM-developed metrics and those used by ChatGPT. Moreover, CSEM can achieve the second-highest scores on the BLEU, COMET-20, and COMET-22 metrics, trailing only the scores obtained from their respective dedicated metrics. This further highlights that CSEM not only competes with but also closely aligns with the performance of well-established metrics. Notably, when compared to ChatGPT, CSEM offers similar advantages but with significantly higher reward query efficiency, as detailed in Section \ref{sec:reward_query_efficiency}. This emphasizes that CSEM not only delivers high-quality rewards but also provides more efficient solutions for sequence evaluation tasks.

\subsubsection{Conclusion}
Despite being tested on various tasks, our CSEM produces consistent results across all of them: 1) With different RL and reranking strategies, all of our CSEM metrics effectively provide rewards that enhance the sequence generation model, further demonstrating the effectiveness of CSEM; 2) We observe that reference-based metrics outperform traditional word-overlap-based metrics such as BLEU, ROUGE, and BLEURT, as they more closely resemble these traditional methods by using references to evaluate the quality of generated sequences; 3) Whether using reference-based or reference-free metrics, the larger versions generally show better performance in RL and reranking; 4) Under traditional evaluation metrics (such as COMET-20, COMET-22, and Chat-based metrics), the RL+RR optimization approach consistently achieves the best performance across various tasks; 5) Compared to traditional metrics, CSEM provides more accurate rewards, highlighting its advantage in evaluation tasks over conventional metrics.

\input{tables/diff_rewards}

\section{Analysis}

\subsection{Impact of the Annotated Data Scale}
For the data collection stage, we explore the impact of the scale of annotated data on our CSEM. Specifically, we use ChatGPT to annotate datasets with a multi-aspect evaluation prompt across different sample scales: \{5K, 10K, 15K, 20K, 25K\}. Based on these annotated datasets, we train the evaluation models. We then compute correlation scores between these evaluation models and ChatGPT to measure the degree of agreement between the evaluation models and ChatGPT. We hypothesize that a higher correlation score with ChatGPT indicates that the evaluation model has learned a stronger evaluation capability from ChatGPT. In this process, we use a test dataset consisting of 1K samples randomly selected from the data collection stage. All test set samples are excluded from the training of the evaluation models. The results are shown in Figure \ref{fig_diff_sample_size_transfer_summary}. They show that CSEM can achieve a significant correlation with ChatGPT using only a small amount of annotated data. Additionally, the performance gain follows a long-tail pattern as the scale of annotated data increases. Similar observations can be made from the results of the summarization task. Considering the cost of LLM annotation, we choose to use 15K LLM-labeled samples for all tasks.

\begin{figure}[!t]
    \centering
    \input{images/diff_sampling_size_summary_style_transfer}
    \vspace{-3mm}
    \caption{
     Correlation scores of the reference-free evaluation models learned employing different sizes of LLM-labeled samples for machine translation and text style transfer tasks.
    }
    \vspace{-2mm}
    \label{fig_diff_sample_size_transfer_summary}
\end{figure}

\begin{figure}
    \centering
    \input{images/LLM-vs-Human-data}
    \caption{Kendall and Spearman correlation scores of evaluation models trained on LLM-labeled data versus human-labeled data. The correlation scores are computed for the coherence evaluation aspect. The symbol $\sharp$ indicates that the evaluation model is trained on human-labeled data.}
    \vspace{-2mm}
    \label{fig:llm-vs-human}
\end{figure}

\subsection{Comparison of LLM-labeled Data with Human-labeled Data for Training Evaluation Models}
To further assess the effectiveness of LLM-labeled data for training evaluation models, we conduct a comparison with human-labeled data. Specifically, we split the SummEval benchmark into a training set of 1K samples and a test set containing the remaining samples. Evaluation models are then trained on this training set. In parallel, we develop several reference-based CSEM-large models using varying amounts of LLM-labeled data: \{1K, 3K, 5K, 7K, 10K\}. The results are presented in Figure \ref{fig:llm-vs-human}, where we also include baselines using BLEURT and COMET trained on human-labeled data. The experiment results indicate that, when the amount of labeled data is comparable, LLM-labeled data does not perform as well as human-labeled data. However, human-labeled datasets are inherently limited in size, while LLM-labeled data can be scaled more easily.

\begin{figure*}
    \centering
    \input{images/analysis_three_images}
    \vspace{-4mm}
    \caption{
    We compare the performance of different sampling methods in sub-figure (a). Here, ``BS'' denotes the beam search. We explore the performance of our CSEM with continuous scores derived from ChatGPT in sub-figure (b). Note that ``+'' indicates the results when using different numbers of source inputs, which are taken from Figure \ref{fig_diff_sample_size_transfer_summary}(a). We use multiple outputs for training evaluation models in sub-figure (c). 
    }
    \label{fig:analysis-sampling-output-continous-score}
\end{figure*}

\subsection{Performance on Different Sampling Methods}
We also conduct experiments using alternative sampling approaches on the IWSLT translation task, including beam search \cite{freitag2017beam}, diverse beam search \cite{roberts2020decoding}, top-$p$ sampling \cite{holtzman2019curious}, and top-$k$ sampling\footnote{Top-$k$ sampling limits the choices of the model to the $k$ most likely next tokens to enhance the coherence of the generated sequence, while top-$p$ (or nucleus) sampling selects tokens cumulatively exceeding a probability threshold $p$, balancing diversity and quality in generated content \cite{holtzman2019curious}.}. Figure \ref{fig:analysis-sampling-output-continous-score}(a) presents the results of computing the correlation scores between the learned evaluation models and ChatGPT. The results indicate that top-$k$ sampling achieves the best performance for the acquired evaluation models. We hypothesize that top-$k$ sampling generates sequences with greater diversity while maintaining high quality compared to other methods, making it better suited for transferring the evaluation capability from LLMs. It is worth noting that although top-$p$ sampling has been shown to generate higher-quality sequences than top-$k$ in \cite{holtzman2019curious}, our experiments reveal that top-$p$ sampling has certain limitations in generating diverse sequences.

\subsection{Using Multiple Outputs for Training Evaluation Models}
Inspired by \cite{freitag2021experts}, we hypothesize that increasing the number of output sequences might improve the performance of the evaluation model learned by CSEM. To test this hypothesis, we train different evaluation models by first sampling 5K source inputs and then increasing the number of output sequences from 1 to 5 for each input. Figure \ref{fig:analysis-sampling-output-continous-score}(c) presents the Spearman correlation scores of the learned evaluation models on the IWSLT dataset. From the figure, we observe that increasing the number of output sequences effectively improves the correlation. Furthermore, we notice that the performance gain from increasing the target outputs (solid line) is slightly smaller than the gain from directly increasing the source inputs (dashed line). Despite this, using multiple target outputs could still be a promising solution when training data is limited.

\subsection{Using Continuous Scores in the Prompt}
We explore the performance of our CSEM using continuous scores derived from ChatGPT. Specifically, following the design in \cite{kocmi2023large}, we modify the segment ``... with one to five stars, where one star means ... and five stars mean ...'' to ``... with a continuous scale from 0 to 100, where a score of zero means ... and a score of one hundred means ...'' in the prompt. We then test the correlation scores with human judgments, as shown in Figure \ref{fig:analysis-sampling-output-continous-score}(b). The results indicate that CSEM with star ratings consistently outperforms CSEM with continuous scores. We attribute this observation to two potential reasons. First, it has been shown in \cite{kocmi2023large} that continuous score-based evaluation for ChatGPT yields inferior performance compared to star-based evaluation. Second, we hypothesize that learning from a continuous score range spanning from 0 to 100 is more challenging than learning from a simplified five-star scale.

\begin{figure}[!t]
    \centering
    \input{images/rating_vs_ranking}
    \vspace{-8mm}
    \caption{
    Comparisons of different ways of data annotation (rating vs. comparison ranking). In sub-figure (a), we present the correlation scores between the evaluation models and human judgments for reference-based coherence evaluation on the SummEval benchmark. In sub-figure (b), we train evaluation models using these two methods and apply them to improve summarization models through RL and reranking.
    }
    \vspace{-2mm}
    \label{fig:rating-vs-comparsion-ranking}
\end{figure}

\subsection{Rating vs. Comparison Ranking}
\label{sec:comparison-rating-ranking}
As described in Section \ref{sec:background-training-evaltaion-models}, the evaluation data annotation typically involves either rating or comparison ranking. Here, we comprehensively compare these two methods in CSEM. Specifically, for comparison ranking, during the data collection stage in CSEM, we select a sequence pair instead of a single sequence. In the data annotation stage, we use the LLM to select the better sequence from the pair rather than assigning scores. The evaluation models are then trained using Eq. \ref{eq:ranking-loss}. In this way, we develop the CSEM-large-ref and CSEM-large metrics for the summarization task. We test CSEM-large-ref on the SummEval benchmark and utilize CSEM-large for RL and reranking, as illustrated in Figure \ref{fig:rating-vs-comparsion-ranking}. From the figure, it is evident that the rating method consistently outperforms comparison ranking. This result is intuitive: comparison ranking primarily focuses on learning a model for ranking purposes, meaning it may not accurately assess the quality of a single sequence but instead evaluates the relative quality between sequences. As a result, in the figure, we observe that the evaluation model trained using comparison ranking performs well in reranking. However, in practical metric applications, the goal is not just ranking; we want to directly assess the quality of a sequence based on a score (e.g., a score of 0 indicates a poor sequence, while a score of 5 indicates a good one). While comparison ranking could achieve similar results, it is not as direct and tends to be less efficient, often requiring a large amount of data to achieve satisfactory performance. In some cases, the model may assign an illogical score, such as giving a sequence a score of 10 (indicating poor quality) for one input and -1 (indicating good quality) for another. In contrast, rating provides a more direct and effective approach. Therefore, in CSEM, we opt for the rating method to train our evaluation models for various tasks.

\input{tables/appendix_classification_loss}

\subsection{Optimization with Classification Loss} \label{sec:classification}
We attempt to train evaluation models with a classification loss implemented using cross-entropy loss, where each star level is regarded as a class.
The optimization objective of the evaluation model is to predict a star class for a sequence:
\begin{eqnarray}
    \mathcal{L}_{\mathrm{class}}&=&-\mathbb{E}_{(x,\hat{y})\sim \mathcal{D}_{r}}\log \mathrm{Pr}_{\phi}(S(x,\hat{y})|x,\hat{y})
\end{eqnarray}
where $S(\cdot)$ represents the star rating class annotated by an LLM, $\mathrm{Pr}_{\phi}(\cdot)$ denotes the probability that the evaluation model $\mathcal{M}$ assigns to labeled rating class, and $\phi$ is the set of parameters of $\mathcal{M}$.
Table \ref{tab:classification} compares correlation scores between the evaluation models, ChatGPT, and human judgments for the reference-based coherence evaluation aspect on the SummEval benchmark. The results indicate that the evaluation model trained with regression loss consistently outperforms the one trained with classification loss. This suggests that regression loss is more effective in CSEM. Moreover, the use of continuous scores offers an advantage in capturing subtle differences between generated sequences, which is also noted in \cite{liu2023gpteval}.

\begin{table}[!t]
    \centering
    \caption{Performance of CSEM with different LLMs on the SummEval benchmark.}
    \scalebox{0.80}{
    \input{tables/diff_LLMs}}
    \label{tab:diffrent-LLMs}
\end{table}

\subsection{Performance on Different LLMs}
Table \ref{tab:diffrent-LLMs} presents the performance of CSEM with different LLMs, including ChatGPT, GPT-4, Llama-3.1-8B-Instruct \cite{grattafiori2024llama}, and Qwen2.5-14B-Instruct \cite{yang2024qwen2}. The experimental results demonstrate that using more powerful LLMs indeed enhances the performance of CSEM. For example, when using GPT-4, we can achieve a Pearson correlation of 0.591 on the coherence evaluation. Moreover, similar to ChatGPT, CSEM can outperform LLMs in many cases, thanks to the specific training methods and model architecture design. This indicates that CSEM does not merely mimic the behaviour of LLMs but also generalizes from the labeled data to achieve stronger performance. However, when using GPT-4, we observe that most of the metrics are lower than the performance of GPT-4 itself. This raises an interesting question: if the performance of an LLM is exceptionally strong, will CSEM struggle to surpass it due to limitations in the underlying encoder (i.e., RoBERTa) capabilities of the model? To address this, we may consider exploring stronger encoders to potentially overcome these limitations and further improve the performance of CSEM.

\begin{table}[!t]
    \centering
    \caption{Performance of CSEM on the general-purpose task.}
    \scalebox{0.9}{
    \input{tables/general_purpose}}
    \label{tab:general-purpose-task}
\end{table}

\subsection{Performance on the General-purpose Task}
We evaluate the performance of CSEM on a general-purpose task using the HelpSteer2 dataset \cite{wang2024helpsteer2}, which is designed for general-purpose evaluation. Specifically, we test on the validation set of HelpSteer2. In this dataset, human annotators annotate scores for each input-output pair across five different dimensions, with scores ranging from 0 to 4. We average these five dimensions to compute the overall quality score for each input-output pair, representing human judgment. Next, we use 15K input samples from the HelpSteer2 training set to develop the CSEM-base and CSEM-large metrics. We then compute the correlation between these two metrics and human judgment on the HelpSteer2 dataset. The results are summarized in Table \ref{tab:general-purpose-task}. Additionally, we test the correlation scores of models such as ChatGPT and Llama-3.1-8B-Instruct for comparison. Furthermore, we compare the performance of several open-source reward models trained using comparison ranking methods, including RM-Gemma-2B, RM-Gemma-7B \cite{dong2023raft}, and Ziya-LLaMA-7B-Reward \cite{ziya-reward-7B}. The results clearly show that, although CSEM uses only 355M parameters, it achieves performance comparable to models with 8B parameters, even surpassing the Llama-3.1-8B-Instruct model. While the reward models are also trained on labeled data, their performance is less satisfactory. This can be attributed to the fact that reward models are focused on learning ranking, as discussed in Section \ref{sec:comparison-rating-ranking}. These experimental results further emphasize the robustness of CSEM, demonstrating its applicability in developing model-based metrics for both specific sequence generation tasks and general-purpose tasks.

\subsection{Comparison of Reward Query Speed} \label{sec:reward_query_efficiency}
We compare the query response time of ChatGPT with our evaluation models developed using the CSEM method. For ChatGPT, each query related to the machine translation task, typically consisting of around 100 words, takes at least 5 seconds or more during peak times due to its large size and the need for API requests. This delay becomes a major obstacle when handling large-scale queries. In contrast, our CSEM-trained evaluation models can process queries in no more than 0.2 seconds. Additionally, these models, with a parameter limit of 355M, can be efficiently deployed locally, greatly enhancing their practical usability. These faster processing times significantly improve the scalability and efficiency of using our CSEM-trained models in real-world scenarios where speed is critical.

\section{Conclusion}
In this work, we have extended the capabilities of LLMs to create labeled data for training evaluation models, addressing the challenge of the scarcity of human-labeled evaluation data in developing model-based metrics. We have introduced the Customized Sequence Evaluation Metric (CSEM) method via LLM-based data annotation. Our extensive experimental results demonstrate the effectiveness of CSEM in developing a wide range of evaluation metrics and highlight its ability to improve sequence generation tasks through RL and reranking techniques significantly.

\section*{Acknowledgments}
This work was supported in part by the National Key R\&D program of China (No. 2022YFE0204900), the National Science Foundation of China (Nos. 62276056 and U24A20334), the Fundamental Research Funds for the Central Universities, the Yunnan Fundamental Research Projects (No. 202401BC070021), and the Program of Introducing Talents of Discipline to Universities, Plan 111 (No. B16009). The authors would like to thank anonymous reviewers for their valuable and helpful comments. 

\bibliographystyle{IEEEtran}
\bibliography{ieee.bib}

\clearpage


\vfill

\end{document}

%% file: use_images/prompt.tex
\begin{figure*}[]
	\centering
	\tikzstyle{every node}=[scale=0.98]
        \begin{tikzpicture}
        \scriptsize{
			\begin{scope}[]
				
				\node [anchor=north,rectangle,rounded corners=5pt,minimum height=1.4in, minimum width=3.3in, line width=1pt, draw=lightgray, dashed] (box) at (0, 0) {};
				
				\node [anchor=north, text width=3.2in] (p1) at ([xshift=0.2em, yshift=-0.2em]box.north) {\textit{\texttt{\sethlcolor{LightBlue}\hl{Based on the human reference}, score the following translation from [}Source Language\texttt{] to [}Target Language\texttt{] with respect to [}Aspect\texttt{] with one to five stars, where one star means [}Description of the Worst Translation on a \sethlcolor{Wheat}\hl{Single Aspect}\texttt{] and five stars mean [}Description of the Perfect Translation on a \sethlcolor{Wheat}\hl{Single Aspect}\texttt{]. Note that [}Definition of the Used \sethlcolor{Wheat}\hl{Single Evaluation Aspect}\texttt{].}}};
				
				\node [anchor=north, text width=3.2in] (p2) at ([xshift=0em, yshift=-0.2em]p1.south) {\texttt{[}\textit{Source Language\texttt{] source: [}Source}\texttt{]}};
				
				\node [anchor=north, text width=3.2in] (p3) at ([xshift=0em, yshift=0.2em]p2.south) {\sethlcolor{LightBlue}\hl{\texttt{[}\textit{Target Language\texttt{] human reference: [}Reference}\texttt{]}}};
				
				\node [anchor=north, text width=3.2in] (p4) at ([xshift=0em, yshift=0.2em]p3.south) {\texttt{[}\textit{Target Language\texttt{] translation: [}Translation}\texttt{]}};
				
				\node [anchor=north, text width=3.2in] (p5) at ([xshift=0em, yshift=0em]p4.south) {\texttt{Stars:}};
				
				\footnotesize{
					\node [anchor=north] (title) at ([xshift=0em, yshift=-0.2em]box.south) {(a) Single-aspect Evaluation for Machine Translation};
				}
			\end{scope}
		}
		
		\scriptsize{
			\begin{scope}[xshift=3.5in]
				
				\node [anchor=north,rectangle,rounded corners=5pt,minimum height=1.4in, minimum width=3.3in, line width=1pt, draw=lightgray, dashed] (box) at (0, 0) {};
				
				\node [anchor=north, text width=3.2in] (p1) at ([xshift=0.2em, yshift=-0.2em]box.north) {\textit{\texttt{\sethlcolor{LightBlue}\hl{Based on the human reference}, score the following translation from [}Source Language\texttt{] to [}Target Language\texttt{] with one to five stars, where one star means [}Description of the Worst Translation on  \sethlcolor{Wheat}\hl{Multiple Aspects}\texttt{] and five stars mean [}Description of the Perfect Translation on \sethlcolor{Wheat}\hl{Multiple Aspects}\texttt{]. Note that [}Definition of the Used \sethlcolor{Wheat}\hl{Multiple Evaluation Aspects}\texttt{].}}};
				
				\node [anchor=north, text width=3.2in] (p2) at ([xshift=0em, yshift=-0.2em]p1.south) {\texttt{[}\textit{Source Language\texttt{] source: [}Source}\texttt{]}};
				
				\node [anchor=north, text width=3.2in] (p3) at ([xshift=0em, yshift=0.2em]p2.south) {\sethlcolor{LightBlue}\hl{\texttt{[}\textit{Target Language\texttt{] human reference: [}Reference}\texttt{]}}};
				
				\node [anchor=north, text width=3.2in] (p4) at ([xshift=0em, yshift=0.2em]p3.south) {\texttt{[}\textit{Target Language\texttt{] translation: [}Translation}\texttt{]}};
				
				\node [anchor=north, text width=3.2in] (p5) at ([xshift=0em, yshift=0em]p4.south) {\texttt{Stars:}};
				
				\footnotesize{
					\node [anchor=north] (title) at ([xshift=0em, yshift=-0.2em]box.south) {(b) Multi-aspect Evaluation for Machine Translation};
				}
			\end{scope}
		}
    \scriptsize{
			\begin{scope}[yshift=-1.7in]
				
				\node [anchor=north,rectangle,rounded corners=5pt,minimum height=1.4in, minimum width=3.3in, line width=1pt, draw=lightgray, dashed] (box) at (0, 0) {};
				
				\node [anchor=north, text width=3.2in] (p1) at ([xshift=0.2em, yshift=-0.2em]box.north) {\textit{\texttt{\sethlcolor{LightBlue}\hl{Based on the human reference}, score the transferred sentence from [}Source Style\texttt{] to [}Target Style\texttt{] with respect to [}Aspect\texttt{] with one to five stars, where one star means [}Description of the Worst Transfer on a \sethlcolor{Wheat}\hl{Single Aspect}\texttt{] and five stars mean [}Description of the Perfect Transfer on a \sethlcolor{Wheat}\hl{Single Aspect}\texttt{]. Note that [}Definition of the Used \sethlcolor{Wheat}\hl{Single Evaluation Aspect}\texttt{].}}};
				
				\node [anchor=north, text width=3.2in] (p2) at ([xshift=0em, yshift=-0.2em]p1.south) {\texttt{Source sentence: [}\textit{Source Sentence}\texttt{]}};
				
				\node [anchor=north, text width=3.2in] (p3) at ([xshift=0em, yshift=0.2em]p2.south) {\sethlcolor{LightBlue}\hl{\texttt{Human reference: [}\textit{Reference}\texttt{]}}};
				
				\node [anchor=north, text width=3.2in] (p4) at ([xshift=0em, yshift=0.2em]p3.south) {\texttt{Transferred sentence: [}\textit{Transferred Sentence}\texttt{]}};
				
				\node [anchor=north, text width=3.2in] (p5) at ([xshift=0em, yshift=0em]p4.south) {\texttt{Stars:}};
				
				\footnotesize{
					\node [anchor=north] (title) at ([xshift=0em, yshift=-0.2em]box.south) {(c) Single-aspect Evaluation for Text Style Transfer};
				}
			\end{scope}
		}

    \scriptsize{
			\begin{scope}[xshift=3.5in, yshift=-1.7in]
				
				\node [anchor=north,rectangle,rounded corners=5pt,minimum height=1.4in, minimum width=3.3in, line width=1pt, draw=lightgray, dashed] (box) at (0, 0) {};
				
				\node [anchor=north, text width=3.2in] (p1) at ([xshift=0.2em, yshift=-0.2em]box.north) {\textit{\texttt{\sethlcolor{LightBlue}\hl{Based on the human reference}, score the transferred sentence from [}Source Style\texttt{] to [}Target Style\texttt{] with one to five stars, where one star means [}Description of the Worst Transfer on a \sethlcolor{Wheat}\hl{Multiple Aspects}\texttt{] and five stars mean [}Description of the Perfect Transfer on \sethlcolor{Wheat}\hl{Multiple Aspects}\texttt{]. Note that [}Definition of the Used \sethlcolor{Wheat}\hl{Multiple Evaluation Aspects}\texttt{].}}};
				
				\node [anchor=north, text width=3.2in] (p2) at ([xshift=0em, yshift=-0.2em]p1.south) {\texttt{Source sentence: [}\textit{Source Sentence}\texttt{]}};
				
				\node [anchor=north, text width=3.2in] (p3) at ([xshift=0em, yshift=0.2em]p2.south) {\sethlcolor{LightBlue}\hl{\texttt{Human reference: [}\textit{Reference}\texttt{]}}};
				
				\node [anchor=north, text width=3.2in] (p4) at ([xshift=0em, yshift=0.2em]p3.south) {\texttt{Transferred sentence: [}\textit{Transferred Sentence}\texttt{]}};
				
				\node [anchor=north, text width=3.2in] (p5) at ([xshift=0em, yshift=0em]p4.south) {\texttt{Stars:}};
				
				\footnotesize{
					\node [anchor=north] (title) at ([xshift=0em, yshift=-0.2em]box.south) {(d) Multi-aspect Evaluation for Text Style Transfer};
				}
			\end{scope}
		}
  	\scriptsize{
			\begin{scope}[yshift=-3.4in]
				
				\node [anchor=north,rectangle,rounded corners=5pt,minimum height=1.4in, minimum width=3.3in, line width=1pt, draw=lightgray, dashed] (box) at (0, 0) {};
				
				\node [anchor=north, text width=3.2in] (p1) at ([xshift=0.2em, yshift=-0.2em]box.north) {\textit{\texttt{\sethlcolor{LightBlue}\hl{Based on the human reference}, score the following summary of the given corresponding article with respect to [}Aspect\texttt{] with one to five stars, where one star means [}Description of the Worst Summary on a \sethlcolor{Wheat}\hl{Single Aspect}\texttt{] and five stars mean [}Description of the Perfect Summary on a \sethlcolor{Wheat}\hl{Single Aspect}\texttt{]. Note that [}Definition of the Used \sethlcolor{Wheat}\hl{Single Evaluation Aspect}\texttt{].}}};
				
				\node [anchor=north, text width=3.2in] (p2) at ([xshift=0em, yshift=-0.2em]p1.south) {\texttt{Article: [}\textit{Article}\texttt{]}};
				
				\node [anchor=north, text width=3.2in] (p3) at ([xshift=0em, yshift=0.2em]p2.south) {\sethlcolor{LightBlue}\hl{\texttt{Human reference: [}\textit{Reference}\texttt{]}}};
				
				\node [anchor=north, text width=3.2in] (p4) at ([xshift=0em, yshift=0.2em]p3.south) {\texttt{Summary: [}\textit{Summary}\texttt{]}};
				
				\node [anchor=north, text width=3.2in] (p5) at ([xshift=0em, yshift=0em]p4.south) {\texttt{Stars:}};
				
				\footnotesize{
					\node [anchor=north] (title) at ([xshift=0em, yshift=-0.2em]box.south) {(e) Single-aspect Evaluation for Summarization};
				}
			\end{scope}
		}
		
		\scriptsize{
			\begin{scope}[xshift=3.5in,yshift=-3.4in]
				
				\node [anchor=north,rectangle,rounded corners=5pt,minimum height=1.4in, minimum width=3.3in, line width=1pt, draw=lightgray, dashed] (box) at (0, 0) {};
				
				\node [anchor=north, text width=3.2in] (p1) at ([xshift=0.2em, yshift=-0.2em]box.north) {\textit{\texttt{\sethlcolor{LightBlue}\hl{Based on the human reference}, score the following summary of the given corresponding article with one to five stars, where one star means [}Description of the Worst Summary on \sethlcolor{Wheat}\hl{Multiple Aspects}\texttt{] and five stars mean [}Description of the Perfect Summary on \sethlcolor{Wheat}\hl{Multiple Aspects}\texttt{]. Note that [}Definition of the Used \sethlcolor{Wheat}\hl{Multiple Evaluation Aspects}\texttt{].}}};
				
				\node [anchor=north, text width=3.2in] (p2) at ([xshift=0em, yshift=-0.2em]p1.south) {\texttt{Article: [}\textit{Article}\texttt{]}};
				
				\node [anchor=north, text width=3.2in] (p3) at ([xshift=0em, yshift=0.2em]p2.south) {\sethlcolor{LightBlue}\hl{\texttt{Human reference: [}\textit{Reference}\texttt{]}}};
				
				\node [anchor=north, text width=3.2in] (p4) at ([xshift=0em, yshift=0.2em]p3.south) {\texttt{Summary: [}\textit{Summary}\texttt{]}};
				
				\node [anchor=north, text width=3.2in] (p5) at ([xshift=0em, yshift=0em]p4.south) {\texttt{Stars:}};
				
				\footnotesize{
					\node [anchor=north] (title) at ([xshift=0em, yshift=-0.2em]box.south) {(f) Multi-aspect Evaluation for Summarization};
				}
			\end{scope}
		}
	\end{tikzpicture}
        \vspace{-2mm}
	\caption{
	    Prompt templates of the single and multi-aspect evaluation.
		Single-aspect evaluation means that the model only needs to consider a specific aspect during the evaluation process.
		Conversely, multi-aspect evaluation means that the model must consider multiple aspects concurrently, e.g., simultaneously considering meaning preservation and grammar to assign stars. Template portions \sethlcolor{LightBlue}\hl{highlighted in blue} denote the reference input used only by reference-based evaluation, and that those \sethlcolor{Wheat}\hl{highlighted in yellow} denote the differentiation between single-aspect evaluation and multi-aspect evaluation.
	}
        \vspace{-4mm}
	\label{fig:prompt}
\end{figure*}

%% file: tables/single_aspect.tex
\begin{table*}[ht]
	\centering
        \caption{The designed single-aspect evaluations and corresponding contents of the unfilled segments for summarization tasks, machine translation, and text style transfer.}
	\scalebox{0.90}{
		\begin{NiceTabular}{l|c|c|l}
			\toprule[1.1pt]
			Evaluation Aspect  &
                \parbox{2.9cm}{\textit{Description of the Worst Task on a Single Aspect}} &
                \parbox{2.9cm}{\textit{Description of the Perfect Task on a Single Aspect}}& 
                \parbox{6.5cm}{\textit{Definition of the Used Single Evaluation Aspect}} \\ \midrule
			\multicolumn{4}{l}{\bf{Machine Translation}}	\\ \midrule
			Accuracy & inaccuracy & perfect accuracy & \parbox{9.8cm}{the accuracy measures whether the translation conveys the intended meaning and information of the source language correctly.}
			\\ \midrule
			Completeness    & incompleteness & perfect completeness &
                \parbox{9.8cm}{the completeness measures whether the translation includes all the information and meaning in the source language.}
			\\ \midrule
			Fluency   & disfluency & perfect fluency & 
                \parbox{9.8cm}{the fluency measures how naturally and smoothly the translation reads in the target language.}
			\\ \midrule
			Style & \parbox{2.9cm}{totally different style with the source sentence}
                & \parbox{2.9cm}{perfect matching with the source sentence}
                & \parbox{9.8cm}{the text style measures how well the translation matches the tone, register, and style of the source language.} \\ \midrule
                \multicolumn{4}{l}{\bf{Text Style Transfer}}							  		\\ \midrule
                Content  & no content preserved & \parbox{2.9cm}{perfect content preservation} & 
                \parbox{9.8cm}{the content preservation measures whether the transferred sentence preserves the meaning and content of the source sentence.}
			\\ \midrule
			Style  & informal style & perfect formal style &
                \parbox{9.8cm}{the style accuracy measures how well the transferred sentence to transfer the style from the source sentence.}
			\\ \midrule
			Fluency  & disfluency & perfect fluency &   
                \parbox{9.8cm}{the fluency measures how naturally and smoothly the transferred sentence reads.}
			\\ \midrule
                \multicolumn{4}{l}{\bf{Summarization}} \\ \midrule
			Coherence  & incoherence & perfect coherence & 
                \parbox{9.8cm}{the coherence measures the quality of all sentences collectively, to the fit together and sound naturally. Consider the quality of the summary as a whole.}
			\\ \midrule
			Consistency  & inconsistency & perfect consistency & 
                \parbox{9.8cm}{the consistency measures whether the facts in the summary are consistent with the facts in the original article. Consider whether the summary does reproduce all facts accurately and does not make up untrue information.}
			\\ \midrule
			Relevance  & irrelevance & perfect relevance &  
                \parbox{9.8cm}{the relevance measures how well the summary captures the key points of the article. Consider whether all and only the important aspects are contained in the summary.}
			\\ \midrule
			Fluency  & disfluency & perfect fluency &   
                \parbox{9.8cm}{the fluency measures the quality of individual sentences, are they well-written and grammatically correct. Consider the quality of individual sentences.}  \\ 
			\bottomrule[1.1pt]
	\end{NiceTabular}}
        \vspace{-3mm}
	\label{tab:single-aspect}
\end{table*}

%% file: tables/multi_aspect.tex
\begin{table*}[ht]
	\centering
        \caption{
		The designed multi-aspect evaluations and corresponding contents of the unfilled segments for machine translation, text style transfer, and summarization tasks.
	}
	\scalebox{0.90}{
	\begin{NiceTabular}{l|c}
		\toprule[1.1pt]
		  Prompt Segment   &  Content 	\\ \midrule
		  \multicolumn{2}{l}{\bf{Machine Translation}}	\\ \midrule
            \parbox{7.0cm}{\textit{Description of the Worst Translation on Multiple Aspects}} & no meaning preserved   \\ \midrule
            \parbox{7.0cm}{\textit{Description of the Perfect Translation on Multiple Aspects}}  & perfect meaning and grammar \\ \midrule
		  \parbox{7.0cm}{\textit{Definition of the Used Multiple Evaluation Aspects}} 
            & \parbox{11.5cm}{the meaning aspect measures whether the translation covers the meaning of the source and the grammar aspect measures the presence of grammatical errors within the translation. } 	\\  \midrule
            \multicolumn{2}{l}{\bf{Text Style Transfer}}   							  		\\ \midrule
		  \parbox{7.0cm}{\textit{Description of the Worst Transfer on Multiple Aspects}} & no meaning preserved   \\ \midrule
            \parbox{7.0cm}{\textit{Description of the Perfect Transfer on Multiple Aspects}}  & formal, perfect meaning and fluency \\ \midrule
		  \parbox{7.0cm}{\textit{Definition of the Used Multiple Evaluation Aspects}} 
            &\parbox{11.5cm}{
	      the formal aspect measures whether the transferred sentence is formal, the meaning aspect measures whether the transferred sentence covers the meaning of the source and    the fluency aspect measures whether the transferred sentence is well-written and grammatically correct. 
		} \\ \midrule
            \multicolumn{2}{l}{\bf{Summarization}}							  		\\ \midrule
            \parbox{7.0cm}{\textit{Description of the Worst Summary on Multiple Aspects}} & no meaning preserved   \\ \midrule
            \parbox{7.0cm}{\textit{Description of the Perfect Summary on Multiple Aspects}}  & including core meaning and fluency \\ \midrule
            \parbox{7.0cm}{\textit{Definition of the Used Multiple Evaluation Aspects}} 
            &\parbox{11.5cm}{
            the meaning aspect measures whether the summary covers the core meaning of the article correctly and the fluency aspect measures whether the summary is well-written and grammatically correct. }                 		
		  \\ 
		\bottomrule[1.1pt]
	\end{NiceTabular}}
        \vspace{-2mm}
	\label{tab:multi-aspect}
\end{table*}

%% file: tables/mt_summary_dataset.tex
\begin{table}[t]
    \centering
    \caption{
        Statistical information on the machine translation and abstractive summarization datasets.
    }
    \scalebox{0.95}{
        \begin{tabular}{c>{\centering\arraybackslash}p{2.5cm}>{\centering\arraybackslash}p{2.5cm}>{\centering\arraybackslash}p{1.5cm}}
        \toprule[1.1pt]
        -             & IWSLT’14 De-En & WMT’14 En-De & CNN/DM \\ \midrule
        Train         &160,239      &3,896,364  &287,113            \\ 
        Valid         &7,283      &39,388   &13,368        \\
        Test          &6,750      &3,003   &11,490          \\ \bottomrule[1.1pt]
        \end{tabular}
    }
\label{tab_statistical_mt_summay_datasets}
\end{table}

%% file: tables/style_transfer_dataset.tex
\begin{table}[t]
    \centering
    \caption{
        Statistical information on the GYAFC dataset.
    }
    \scalebox{0.95}{
        \begin{tabular}{l>{\centering\arraybackslash}p{1.5cm}>{\centering\arraybackslash}p{1.5cm}>{\centering\arraybackslash}p{1.5cm}}
        \toprule[1.1pt]
        \multicolumn{1}{c}{\multirow{2}{*}{Domain}} & \multicolumn{3}{c}{Informal $\rightarrow$ Formal} \\ \cmidrule(l){2-4}  
        \multicolumn{1}{c}{}                        & Train           & Valid          & Test           \\ \midrule
        F\&R                                        & 51,967          & 2,788          & 1,332          \\
        E\&M                                        & 52,595          & 2,877          & 1,416    \\
        \bottomrule[1.1pt]
        \end{tabular}
    }
    \vspace{-4mm}
\label{tab_statistical_style_tranfer_datasets}
\end{table}

%% file: tables/summary_multi_aspect.tex
\begin{table*}[]
    \centering
    \caption{
    Performance of the learned reference-based and reference-free evaluation models on the SummEval benchmark.
    }
    \label{tab:summary-multi-aspect}
    \scalebox{0.88}{
    \begin{tabular}{lcccccccccccccccc}
    \toprule[1.1pt]
    \multirow{2}{*}{Method}& \multirow{2}{*}{\#Params} & \multicolumn{3}{c}{Coherence} & \multicolumn{3}{c}{Relevance} & \multicolumn{3}{c}{Consistency} & \multicolumn{3}{c}{Fluency}   & \multicolumn{3}{c}{Avg.}  \\ \cmidrule(l){3-5} \cmidrule(l){6-8} \cmidrule(l){9-11} \cmidrule(l){12-14} \cmidrule(l){15-17}
    & & Kend. & Spear. & Pear. & Kend. & Spear. & Pear. & Kend. & Spear. & Pear. & Kend. & Spear. & Pear. & Kend. & Spear. & Pear. \\ \midrule
    \multicolumn{17}{l}{\textit{\textbf{Reference-based Evaluation}}} \\ \midrule
    ChatGPT                  & -&0.407 & 0.474  & 0.491 & 0.378 & 0.430  & 0.457 & 0.375 & 0.403  & 0.489 & 0.319 & 0.339  & 0.409 & 0.370 & 0.411 & 0.461      \\ \hdashline
    ROUGE-1                  & -& 0.126 & 0.167  & 0.160 & 0.252 & 0.326  & 0.359 & 0.130 & 0.160  & 0.224 & 0.094 & 0.115 & 0.158 & 0.150 & 0.192 & 0.225        \\
    ROUGE-2                  & -& 0.139 & 0.184  & 0.174 & 0.219 & 0.290  & 0.327 & 0.155 & 0.187  & 0.246 & 0.128 & 0.159 & 0.185 & 0.160 & 0.205 & 0.233        \\
    ROUGE-L                  & -& 0.099 & 0.128  & 0.102 & 0.237 & 0.311  & 0.342 & 0.092 & 0.115  & 0.189 & 0.084 & 0.105 & 0.141 & 0.128 & 0.165 & 0.194          \\
    BERTScore                & 110M& 0.211 & 0.283  & 0.310 & 0.243 & 0.311  & 0.346 & 0.090 & 0.110  & 0.152 & 0.158 & 0.192 & 0.209 & 0.175 & 0.224 & 0.254          \\
    MoverScore               & 110M& 0.118 & 0.159  & 0.167 & 0.244 & 0.318  & 0.371 & 0.127 & 0.157  & 0.224 & 0.105 & 0.129 & 0.176 & 0.148 & 0.191 & 0.234          \\
    BARTScore                & 406M&0.250 & 0.322  & 0.345 & 0.197 & 0.264  & 0.290 & 0.256 & 0.311  & 0.321 & 0.203 & 0.248 & 0.260 & 0.227 & 0.286 & 0.304          \\
    BARTScore-FT & 406M& 0.342 & 0.448  & 0.458 & 0.273 & 0.356  & 0.369 & 0.315 & 0.382  & 0.422 & \bf0.292 & \bf0.356 & \bf0.407 & 0.305 & 0.385 & 0.414  \\
    CSEM-base (Ours)    & 125M &0.372 & 0.512  & 0.535 & 0.356 & 0.487  & 0.517 & 0.364 & 0.423  & 0.502 & 0.252 & 0.291  & 0.372 & 0.336 & 0.428 & 0.482      \\
    CSEM-large (Ours)   & 355M&\bf0.386 & \bf0.529  & \bf0.566 & \bf0.365 & \bf0.497  & \bf0.531 & \bf0.382 & \bf0.431  &\bf0.513 & 0.275 & 0.312  & 0.393 & \bf0.352 & \bf0.442 & \bf0.501     \\ \midrule
    \multicolumn{17}{l}{\textit{\textbf{Reference-free Evaluation}}} \\ \midrule
    ChatGPT                  & -   &0.403 & 0.470 & 0.484 & 0.374 & 0.428  & 0.454 & 0.389 & 0.419  & 0.517 & 0.329 & 0.353 & 0.415 & 0.374 & 0.417 & 0.468    \\  \hdashline
    CSEM-base (Ours)    & 125M&\bf{0.336} & \bf{0.467} & \bf{0.481} & \bf{0.324} & \bf{0.446}  & 0.461 & 0.370 & 0.429  & 0.508 & 0.196 & 0.252 & 0.326 & \bf{0.306} & \bf{0.399} & 0.444   \\
    CSEM-large (Ours)         & 355M &0.307 & 0.428 & 0.461 & 0.322 & 0.442  & \bf0.474 & \bf{0.372} & \bf0.433  & \bf0.524 & \bf{0.198} & \bf{0.254} & \bf{0.342} & 0.300 & 0.389 & \bf{0.450}  \\ 
    \toprule[1.1pt]
    \multicolumn{17}{l}{\parbox{20cm}{All evaluation models are single-aspect evaluation models, focusing on coherence, relevance, consistency, and fluency aspects, respectively. BARTScore-FT is an improved version of BARTScore based on the BART-large fine-tuned with the CNN/DM dataset \cite{yuan2021bartscore}. Given that we conducted experiments with the same prompts, hyper-parameters, and version of the ChatGPT model as \cite{wang2023chatgpt}, the results of ChatGPT are taken from this work. 
    The best result for each group is \textbf{bolded}. Note that we do not include comparisons with ChatGPT, as it is utilized to annotate the data.
    “Avg.” denotes the average performance.
    “Kend.”, “Spear.”, and “Pear.” denote Sample-level Kendall-Tau, Spearman, and Pearson correlation scores, respectively.
    The suffix “-base” and “-large” denote that we employ RoBERTa-base and RoBERTa-large in our evaluation model architecture.}} \\
    \end{tabular}
    }
\end{table*}

%% file: tables/mt_res.tex
\begin{table*}[]
    \centering
    \caption{Results on the machine translation task.}
    \scalebox{0.88}{
        \begin{tabular}{l>{\centering\arraybackslash}p{1.3cm}>{\centering\arraybackslash}p{1.4cm}>{\centering\arraybackslash}p{1.4cm}>{\centering\arraybackslash}p{1.3cm}>{\centering\arraybackslash}p{1.2cm}>{\centering\arraybackslash}p{1.2cm}>{\centering\arraybackslash}p{1.4cm}>{\centering\arraybackslash}p{1.4cm}>{\centering\arraybackslash}p{1.2cm}>{\centering\arraybackslash}p{1.2cm}}
        \toprule[1.1pt]
        \multirow{2}{*}{Method} & \multicolumn{5}{c}{IWSLT’14 De-En}    & \multicolumn{5}{c}{WMT’14 En-De}\\ \cmidrule(l){2-6}  \cmidrule(l){7-11}                 
                                & BLEU  & COMET-20 & COMET-22 & Chat. & ChatRef. &BLEU  & COMET-20 & COMET-22 & Chat. & ChatRef.\\ \midrule
        MLE               & 34.57 & 37.07  &79.32& 75.03                & 65.61  &27.26  &49.79 &83.36 &91.32 & 68.13                 \\ \midrule
        \multicolumn{11}{l}{\textit{\textbf{with one multi-aspect evaluation model}}}                                 \\ \midrule
        RL w/ CSEM-base-ref&    \bf{34.34}&     38.24&  79.61&  75.58&  66.20&  \bf{27.03}&     51.10&  83.64&  91.85&  69.14   \\
        RL w/ CSEM-base&        33.35&  37.82&  79.46&  75.46&  66.10&  26.31&  50.75&  83.54&  92.07&  69.12   \\
        RL+RR w/ CSEM-base&     28.12&  \bf{40.20}&     \bf{80.16}&     \bf{75.97}&     \bf{66.69}&     20.63&  \bf{52.35}&     \bf{84.56}&     \bf{93.13}&     \bf{70.12}      \\ \midrule
        RL w/ CSEM-large-ref&   \bf{34.73}&     38.06&  79.55&  75.76&  66.16&  \bf{27.34}&     51.35&  83.96&  91.83&  69.38   \\
        RL w/ CSEM-large&       33.46&  38.02&  79.56&  75.64&  66.23&  26.00&  51.01&  83.67&  92.12&  69.26   \\
        RL+RR w/ CSEM-large&    28.97&  \bf{44.53}&     \bf{81.71}&     \bf{78.53}&     \bf{68.19}&     21.65&  \bf{52.97}&     \bf{85.25}&     \bf{93.55}&     \bf{70.83}      \\ \midrule
        \multicolumn{11}{l}{\textit{\textbf{with multiple single-aspect evaluation models}}}                                 \\ \midrule
        RR w/ CSEM-base&        28.03&  40.37&  80.17&  75.34&  66.58&  18.96&  51.12&  84.12&  92.59&  69.83   \\
        RR w/ CSEM-large&       \bf{28.90}&     \bf{44.37}&     \bf{81.61}&     \bf{78.32}&     \bf{67.70}&     \bf{20.72}&     \bf{51.74}&     \bf{84.67}&     \bf{93.21}&     \bf{70.57}      \\
        \toprule[1.1pt]
        \multicolumn{11}{l}{\parbox{20cm}{RR denotes the reranking approach.
        The performance of the pre-trained pre-trained generative model is shown in the “MLE” row.
        We report some commonly used evaluation metrics on test sets, including BLEU \cite{papineni2002bleu}, COMET-20 \cite{rei2020unbabel}, and COMET-22 \cite{rei2022comet}.
        We also report the multi-aspect evaluation scores obtained from ChatGPT using reference-based (ChatRef.) and reference-free (Chat.) evaluations. ChatGPT scores are scaled by min-max normalization and multiplied by 100, resulting in a range from 0 to 100. We propose integrating RL with reranking (RL+RR) to enhance the optimization against our evaluation models. Specifically, we use RL to optimize a multi-aspect and reference-free evaluation model and then rerank using this evaluation model. 
        }}
        \end{tabular} 
    }
    \vspace{-4mm}
    \label{tab:mt-res}
\end{table*}

%% file: tables/style_trans_res.tex
\begin{table*}[t]
    \centering
    \caption{Results on the text style transfer task.}
    \scalebox{0.88}{
        \begin{tabular}{l>{\centering\arraybackslash}p{1.3cm}>{\centering\arraybackslash}p{1.4cm}>{\centering\arraybackslash}p{1.4cm}>{\centering\arraybackslash}p{1.3cm}>{\centering\arraybackslash}p{1.2cm}>{\centering\arraybackslash}p{1.2cm}>{\centering\arraybackslash}p{1.4cm}>{\centering\arraybackslash}p{1.4cm}>{\centering\arraybackslash}p{1.2cm}>{\centering\arraybackslash}p{1.2cm}}
        \toprule[1.1pt]
        \multirow{2}{*}{Method} & \multicolumn{5}{c}{E\&M Domain}    & \multicolumn{5}{c}{F\&R Domain}\\ \cmidrule(l){2-6}  \cmidrule(l){7-11}                 
                                & Accuracy  & BLEU & BLEURT & Chat. & ChatRef.  &Accuracy  & BLEU & BLEURT & Chat. & ChatRef.\\ \midrule
        MLE                    & 84.46     &71.98   &73.99      & 56.18                 & 62.46 & 85.96 & 73.97 & 73.33   &58.01             &62.89 \\ \midrule
        \multicolumn{11}{l}{\textit{\textbf{with one multi-aspect evaluation model}}}                                 \\ \midrule
        RL w/ CSEM-base-ref&    \bf{86.16}&     \bf{70.85}&     \bf{74.35}&     57.12&  64.04&  \bf{86.01}&     \bf{73.90}&     73.47&  58.07&  63.16   \\
        RL w/ CSEM-base&        83.76&  70.36&  74.11&  56.47&  63.70&  85.44&  73.87&  \bf{73.50}&     58.84&  63.25   \\
        RL+RR w/ CSEM-base&     74.29&  64.42&  73.50&  \bf{57.38}&     \bf{64.27}&     75.30&  68.18&  73.18&  \bf{59.14}&     \bf{63.68}      \\ \midrule
        RL w/ CSEM-large-ref&   \bf{86.09}&     \bf{70.87}&     74.35&  56.83&  64.15&  86.89&  \bf{74.05}&     73.52&  58.34&  63.63   \\
        RL w/ CSEM-large&       85.95&  70.63&  \bf{74.36}&     56.90&  64.02&  \bf{87.01}&     73.88&  73.22&  59.30&  63.51   \\
        RL+RR w/ CSEM-large&    74.44&  66.15&  73.83&  \bf{57.87}&     \bf{64.64}&     77.70&  69.52&  \bf{73.69}&     \bf{59.32}&     \bf{63.96}      \\ \midrule
        \multicolumn{11}{l}{\textit{\textbf{with multiple single-aspect evaluation models}}}                                 \\ \midrule
        RR w/ CSEM-base&        75.14&  65.47&  73.48&  57.28&  64.18&  76.50&  \bf{68.58}&     73.01&  58.29&  63.20   \\
        RR w/ CSEM-large&       \bf{79.03}&     \bf{65.68}&     \bf{73.77}&     \bf{57.56}&     \bf{64.52}&     \bf{78.75}&     68.31&  \bf{73.20}&     \bf{58.73}&     \bf{63.51}      \\
        \toprule[1.1pt]
        \multicolumn{11}{l}{\parbox{20cm}{ We conduct experiments with informal-to-formal style transfer on E\&M and F\&R domains.
        We report Accuracy, BLEU, and BLEURT \cite{sellam2020bleurt} on test sets.
        Accuracy denotes the accuracy of the output labeled as the target style by a binary classifier obtained from \cite{lai2021thank}.}}
        \end{tabular}
    }
    \vspace{-2mm}
    \label{tab:style-transfer-res}
\end{table*}

%% file: tables/summary_res.tex
\begin{table}[]
    \centering
    \caption{
        Results on the summarization task.
    }
    \scalebox{0.88}{
        \begin{tabular}{l>{\centering\arraybackslash}p{0.88cm}>{\centering\arraybackslash}p{0.88cm}>{\centering\arraybackslash}p{0.88cm}>{\centering\arraybackslash}p{0.88cm}>{\centering\arraybackslash}p{0.88cm}}
        \toprule[1.1pt]
        Method  & RG-1  & RG-2 & RG-L & Chat. & ChatRef. \\ \midrule
        MLE               & 40.26 & 17.85  & 37.06 & 73.68                                        & 72.29                  \\ \midrule
        \multicolumn{6}{l}{\textit{\textbf{with one multi-aspect evaluation model}}}                                 \\ \midrule
        RL w/ CSEM-base-ref&    \bf{40.32}&     \bf{17.86}&     \bf{37.11}&     73.79&  75.48   \\
        RL w/ CSEM-base&        40.22&  17.83&  37.02&  73.75&  75.03   \\
        RL+RR w/ CSEM-base&     38.67&  15.50&  35.53&  \bf{73.96}&     \bf{75.52}      \\  \midrule
        RL w/ CSEM-large-ref&   \bf{40.49}&     \bf{17.91}&     \bf{37.18}&     73.77&  75.45   \\
        RL w/ CSEM-large&       40.31&  17.88&  37.11&  73.84&  75.12   \\
        RL+RR w/ CSEM-large&    39.04&  15.83&  35.95&  \bf{74.42}&     \bf{75.78}              \\ \midrule
        \multicolumn{6}{l}{\textit{\textbf{with multiple single-aspect evaluation models}}}                                 \\ \midrule
        RR w/ CSEM-base&        \bf{39.62}&     15.49&  35.51&  73.78&  75.15   \\
        RR w/ CSEM-large&       39.09&  \bf{15.88}&     \bf{35.99}&     \bf{74.01}&     \bf{75.65}  \\
        \bottomrule[1.1pt]
        \end{tabular}
    }
    \label{tab:summary-res}
    \vspace{-1mm}
\end{table}

%% file: tables/case_study_IWSLT_WMT.tex
\begin{table*}[!t]
    \centering
    \caption{Several examples from the machine translation task on the IWSLT'14 De-En and WMT'14 En-De test sets.}
    \scalebox{0.90}{
    \begin{tabular}{cll}
    \toprule[1.1pt]
    \multicolumn{3}{c}{\textbf{\textit{Cases on the IWSLT'14 De-En Dataset}}} \\ \midrule
    \multirow{4}{*}{\begin{tabular}[c]{@{}c@{}}Case 1 \\ (Meaning Error)\end{tabular}}    
    & Source    & a hat diese vorteile und \textbf{risiken} . \\
    & Reference & a has these benefits , and \textbf{\color{teal}{these risks}} . \\ \cmidrule{2-3} 
    & MLE       & a has these benefits and \textbf{\color{red}{benefits}} . \\
    & Ours      & a has these advantages and \textbf{\color{teal}{these risks}} . \\ \midrule
\multirow{4}{*}{\begin{tabular}[c]{@{}c@{}}Case 2 \\ (Meaning Incompleteness)\end{tabular}}      
    & Source    & \parbox{14.5cm}{wir stoßen hier in erster linie an die grenzen der biologischen realität und \textbf{an jene unserer vorstellung} .} \\
    & Reference & \parbox{14.5cm}{we 're limited here primarily by a biological reality and \textbf{\color{teal}{our imagination}} .} \\ \cmidrule{2-3}
    & MLE       & \parbox{14.5cm}{we are first bumping the limits of biological reality and \textbf{\color{red}{those of us who are}} .} \\
    & Ours      & \parbox{14.5cm}{we 're starting to push the limits of biological reality and \textbf{\color{teal}{that of our imagination}} .} \\ \midrule
    \multirow{4}{*}{\begin{tabular}[c]{@{}c@{}}Case 3 \\ (Grammar Error)\end{tabular}}     
    & Source    & \parbox{14.5cm}{von sokrates ist bekannt , dass er daran glaubte , er hätte einen dämon , der ihm weisheiten aus weiter ferne mitteilte .} \\
    & Reference & \parbox{14.5cm}{socrates , famously , believed that he had a daemon who spoke wisdom to him from afar .} \\ \cmidrule{2-3} 
    & MLE       & \parbox{14.5cm}{he 's known as socrates , \textbf{\color{red}{that}} he believed he had a demonstration that would tell \textbf{\color{red}{him refused}} far away .} \\
    & Ours      & \parbox{14.5cm}{from socrates , he was known to believe he had a blip that shared wisdom for him from far away .} \\  \midrule
    
    \multicolumn{3}{c}{\textbf{\textit{Cases on the WMT'14 En-De Dataset}}} \\ \midrule
    \multirow{4}{*}{\begin{tabular}[c]{@{}c@{}}Case 1 \\ (Meaning Error)\end{tabular}}    
    & Source    & \parbox{14.5cm}{If the pedestrian presses \textbf{the button at the traffic lights} , the top radar sensor checks the traffic status .} \\
    & Reference & \parbox{14.5cm}{Drückt der Fußgänger \textbf{\color{teal}{den Ampelknopf}} , testet der obere Radarsensor die Verkehrslage .} \\ \cmidrule{2-3} 
    & MLE       & \parbox{14.5cm}{Wenn der Fußgänger den Knopf an \textbf{\color{red}{den Verkehrsampeln}} drückt , prüft der obere Radar-Sensor den Verkehrszustand .} \\
    & Ours      & \parbox{14.5cm}{Wenn der Fußgänger \textbf{\color{teal}{den Knopf an der Ampel}} drückt , prüft der obere Radar-Sensor den Verkehrszustand .} \\ \midrule
    
    \multirow{4}{*}{\begin{tabular}[c]{@{}c@{}}Case 2 \\ (Meaning Incompleteness)\end{tabular}}      
    & Source    & \parbox{14.5cm}{The free marketeers at the Reason Foundation are also \textbf{fond of} having drivers pay per mile .} \\
    & Reference & \parbox{14.5cm}{Auch die freien Vermarkter der Reason Foundation sind von der Idee \textbf{\textcolor{teal}{angetan}} , Fahrer nach zurückgelegter Strecke zahlen zu lassen .} \\ \cmidrule{2-3}
    & MLE       & \parbox{14.5cm}{Die Freimarktwirtschaftler der Reason Foundation sind auch \textbf{\textcolor{red}{dafür}} , dass Autofahrer pro Meile bezahlen .} \\
    & Ours      & \parbox{14.5cm}{Die Anhänger der freien Marktwirtschaft bei der Reason Foundation \textbf{\textcolor{teal}{angetan}} ebenfalls , dass Autofahrer pro gefahrene Meile bezahlen .} \\ \midrule
    
    \multirow{4}{*}{\begin{tabular}[c]{@{}c@{}}Case 3 \\ (Grammar Error)\end{tabular}}     
    & Source    & \parbox{14.5cm}{The firm was not originally in the business of helping states tax drivers .} \\
    & Reference & \parbox{14.5cm}{Die Firma ist ursprünglich nicht angetreten , um Bundesstaaten bei der Besteuerung von Autofahrern zu helfen .} \\ \cmidrule{2-3} 
    & MLE       & \parbox{14.5cm}{die Firma war ursprünglich nicht im Geschäft, den Staaten \textbf{\textcolor{red}{dabei zu helfen}} , Fahrer zu besteuern .} \\
    & Ours      & \parbox{14.5cm}{Das Unternehmen war ursprünglich nicht im Geschäft , den Staaten bei der Besteuerung von Fahrern zu helfen .} \\
    \toprule[1.1pt]
    
    \multicolumn{3}{l}{\parbox{16.8cm}{\textbf{\color{teal}{Green}} words are good translations, while \textbf{\color{red}{Red}} words are bad translations.}}
    \end{tabular}}
    \vspace{-3mm}
    \label{tab:case-study-iwslt-wmt}
\end{table*}

%% file: tables/human_eval_res.tex
\begin{table}[!t]
\centering
\caption{
    Results for human evaluation on the machine translation.
}
\scalebox{0.88}{
\begin{tabular}{lc}
    \toprule[1.1pt]
        Method & Human R.          \\ \midrule
        MLE                        &          3.755            \\ \midrule
        \multicolumn{2}{l}{\textit{\textbf{with one multi-aspect evaluation model}}}   \\ \midrule
        RL w/ CSEM-base-ref    & 3.814\\
        RL w/ CSEM-base        & 3.805\\
        RL+RR w/ CSEM-base     & \bf3.862 \\  \midrule
        RL w/ CSEM-large-ref \hspace{1cm}   & 3.830 \\
        RL w/ CSEM-large       & 3.825 \\
        RL+RR w/ CSEM-large    & \bf3.916  \\  \midrule
        \multicolumn{2}{l}{\textit{\textbf{with multiple single-aspect evaluation models \hspace{1cm}}}} \\ \midrule
        RR w/ CSEM-base        & 3.873  \\
        RR w/ CSEM-large       & \bf3.930   \\ 
    \bottomrule[1.1pt]
\end{tabular}}
\vspace{-2mm}
\label{tab_human_eval}
\end{table}

%% file: tables/diff_rewards.tex
\begin{table}[!t]
\centering
\caption{
    Employing RL to train translation models with different reference-based rewards on the IWSLT’14 task.
}
\scalebox{0.83}{
\begin{tabular}{l>{\centering\arraybackslash}p{1cm}>{\centering\arraybackslash}p{1.4cm}>{\centering\arraybackslash}p{1.4cm}>{\centering\arraybackslash}p{1.3cm}>{\centering\arraybackslash}p{1.2cm}}
    \toprule[1.1pt]
        Reward              & BLEU & COMET-20 & COMET-22 & ChatRef. & Human R. \\ \midrule
        ChatGPT             & 34.12 & 38.33 & 79.69 & 66.41 & 3.87  \\ \midrule
        CSEM (Ours)&    34.73&  38.06&  79.55&  \bf{66.16}&     \bf{3.83}       \\
        BLEU&   \bf{34.91}&     37.67&  78.40&  64.80&  3.69    \\
        COMET-20&       32.29&  \bf{38.52}&     79.42&  65.28&  3.75    \\
        COMET-22&       32.78&  37.93&  \bf{79.65}&     65.37&  3.78    \\
    \toprule[1.1pt]
    \multicolumn{6}{l}{\parbox{10cm}{
        We report the results of RL with CSEM-large-ref in the “CSEM” row.
        We also report the human evaluation results for each translation model as described in Section \ref{sec:human_evaluation}.
    }}
\end{tabular}}
\vspace{-2mm}
\label{tab:diff_rewards}
\end{table}

%% file: images/diff_sampling_size_summary_style_transfer.tex
\centering
\begin{tabular}{@{}c@{\hspace{-3.1cm}}c@{}}
\centering
\begin{tikzpicture}[scale=0.24]
\matrix (m) [
        fill=white,
        draw=black,
        at={(3.8, 13)},
        anchor=north west,
        cells={anchor=west},
        scale=0.8,
        inner sep=1.8pt]
        {
            &[1.0em] \LegendImage{color=mygreen,mark=triangle*,mark size=1.5pt} &  \LegendEntry{\scriptsize Kend.}; \
            &[1em] \LegendImage{color=myred,mark=pentagon*,mark size=1.5pt} &  \LegendEntry{\scriptsize Spear.};&[1em] \
            &[1em] \LegendImage{color=myblue,mark=diamond*,mark size=1.5pt} &  \LegendEntry{\scriptsize Pear.}; &[1em]\\
        };	
\begin{axis}[
        ymajorgrids,xmajorgrids,
        grid style=dashed,
        width=0.9\textwidth,
        height=.65\textwidth,
        symbolic x coords={5K, 10K, 15K, 20K, 25K},
        xtick=data,
        x tick label style={scale=2.4},
        y tick label style={scale=2.4,
                            /pgf/number format/fixed,
                            /pgf/number format/fixed zerofill,
                            /pgf/number format/precision=2},
        ymin=0.45, ymax=0.75,
        ytick={0.45,0.50,...,0.75},
        xlabel={Sample Size},
        ylabel={Correlation Score},
        ylabel style={yshift=3.5em, scale=2.5},
        xlabel style={yshift=-1em, scale=2.5},]
    \addplot [mygreen, mark=triangle*, mark size = 6, line width=1.5pt] coordinates { 
    (5K, 0.498)
    (10K, 0.530)
    (15K, 0.544)
    (20K, 0.549)
    (25K, 0.546)};

    \addplot [myred, mark=pentagon*, mark size = 6, line width=1.5pt] coordinates { 
    (5K, 0.628)
    (10K, 0.661)
    (15K, 0.676)
    (20K, 0.680)
    (25K, 0.677)};

    \addplot [myblue, mark=diamond*, mark size = 6, line width=1.5pt] coordinates { 
    (5K, 0.691)
    (10K, 0.704)
    (15K, 0.718)
    (20K, 0.724)
    (25K, 0.723)};
\end{axis}
\node [xshift=5em, yshift=-4ex] {\footnotesize (a) Machine Translation};
\end{tikzpicture}
&
\centering
\begin{tikzpicture}[scale=0.24]
    \begin{axis}[
            ymajorgrids,xmajorgrids,
            grid style=dashed,
            width=0.9\textwidth,
            height=.65\textwidth,
            symbolic x coords={5K, 10K, 15K, 20K, 25K},
            xtick=data,
            x tick label style={scale=2.4},
            y tick label style={scale=2.4,
                                /pgf/number format/fixed,
                                /pgf/number format/fixed zerofill,
                                /pgf/number format/precision=2},
            ymin=0.45, ymax=0.65,
            ytick={0.45,0.50,...,0.65},
            xlabel={Sample Size},
            ylabel={Correlation Score},
            ylabel style={yshift=3.5em, scale=2.5},
            xlabel style={yshift=-1em, scale=2.5},]
        \addplot [mygreen, mark=triangle*, mark size = 6, line width=1.5pt] coordinates { 
        (5K, 0.486)
        (10K, 0.491)
        (15K, 0.503)
        (20K, 0.511)
        (25K, 0.517)};

        \addplot [myred, mark=pentagon*, mark size = 6, line width=1.5pt] coordinates { 
        (5K, 0.590)
        (10K, 0.598)
        (15K, 0.609)
        (20K, 0.624)
        (25K, 0.628)};

        \addplot [myblue, mark=diamond*, mark size = 6, line width=1.5pt] coordinates { 
        (5K, 0.605)
        (10K, 0.609)
        (15K, 0.623)
        (20K, 0.630)
        (25K, 0.637)};

    \end{axis}
    \node [xshift=5em, yshift=-4ex] {\footnotesize (b) Text Style Transfer};
\end{tikzpicture}
\end{tabular}

%% file: images/LLM-vs-Human-data.tex
\centering
\begin{tabular}{@{}c@{\hspace{-3.1cm}}c@{}}
\centering
\begin{tikzpicture}[scale=0.24]
\matrix (m) [
        fill=white,
        draw=black,
        at={(3.8, 13)},
        anchor=north west,
        cells={anchor=west},
        scale=0.8,
        inner sep=1.4pt]
        {
            &[0em] \LegendImage{color=mygreen,mark size=1.5pt} &  \LegendEntry{\scriptsize BLEURT$^\sharp$}; \
            &[0em] \LegendImage{color=myred,mark size=1.5pt} &  \LegendEntry{\scriptsize COMET$^\sharp$};&[1em] \
            &[0em] \LegendImage{color=myblue,mark=diamond*,mark size=1.5pt} &  \LegendEntry{\scriptsize CSEM}; &[1em]\\
        };	
\begin{axis}[
        ymajorgrids,xmajorgrids,
        grid style=dashed,
        width=0.9\textwidth,
        height=.65\textwidth,
        symbolic x coords={1K, 3K, 5K, 7K, 10K},
        xtick=data,
        x tick label style={scale=2.4},
        y tick label style={scale=2.4,
                            /pgf/number format/fixed,
                            /pgf/number format/fixed zerofill,
                            /pgf/number format/precision=2},
        ymin=0.23, ymax=0.37,
        ytick={0.24,0.26,...,0.36},
        xlabel={Size of LLM-labeled Data},
        ylabel={Kend.},
        ylabel style={yshift=3.5em, scale=2.5},
        xlabel style={yshift=-1em, scale=2.5},]
        \addplot [mygreen, mark size = 6, line width=1.5pt] coordinates { 
        (1K, 0.287)
        (3K, 0.287)
        (5K, 0.287)
        (7K, 0.287)
        (10K, 0.287)};

        \addplot [myred, mark size = 6, line width=1.5pt] coordinates { 
        (1K, 0.312)
        (3K, 0.312)
        (5K, 0.312)
        (7K, 0.312)
        (10K, 0.312)};

        \addplot [myblue, mark=diamond*, mark size = 6, line width=1.5pt] coordinates { 
        (1K, 0.261)
        (3K, 0.295)
        (5K, 0.331)
        (7K, 0.342)
        (10K, 0.350)};
\end{axis}
\end{tikzpicture}
&
\centering
\begin{tikzpicture}[scale=0.24]
    \begin{axis}[
            ymajorgrids,xmajorgrids,
            grid style=dashed,
            width=0.9\textwidth,
            height=.65\textwidth,
            symbolic x coords={1K, 3K, 5K, 7K, 10K},
            xtick=data,
            x tick label style={scale=2.4},
            y tick label style={scale=2.4,
                                /pgf/number format/fixed,
                                /pgf/number format/fixed zerofill,
                                /pgf/number format/precision=2},
            ymin=0.35, ymax=0.49,
            ytick={0.36,0.38,...,0.48},
            xlabel={Size of LLM-labeled Data},
            ylabel={Spear.},
            ylabel style={yshift=3.5em, scale=2.5},
            xlabel style={yshift=-1em, scale=2.5},]
        \addplot [mygreen, mark size = 6, line width=1.5pt] coordinates { 
        (1K, 0.388)
        (3K, 0.388)
        (5K, 0.388)
        (7K, 0.388)
        (10K, 0.388)};
        
        \addplot [myred, mark size = 6, line width=1.5pt] coordinates { 
        (1K, 0.423)
        (3K, 0.423)
        (5K, 0.423)
        (7K, 0.423)
        (10K, 0.423)};

        \addplot [myblue, mark=diamond*, mark size = 6, line width=1.5pt] coordinates { 
        (1K, 0.376)
        (3K, 0.414)
        (5K, 0.457)
        (7K, 0.468)
        (10K, 0.475)};

    \end{axis}
\end{tikzpicture}
\end{tabular}

%% file: images/analysis_three_images.tex
\pgfplotsset{compat=1.18}

\begin{tikzpicture}
  \begin{scope}
    \begin{axis}[
      ymajorgrids,
      grid style=dashed,
      legend style={at={(0.5,1.05)}, anchor=south, font=\scriptsize,/tikz/every even column/.append style={column sep=.1cm}},
      legend cell align={left},
      legend columns=2,
      ybar,
      enlarge x limits=0.5,
      xtick align=inside,
      height=.28\textwidth,
      width=.33\textwidth,
      ylabel={Correlation Score},
      symbolic x coords={0, 1},
      xtick=data,
      nodes near coords align={vertical},
      xticklabels={Spear., Pear.},
      x tick label style={font=\scriptsize},
      ymin=0.45, ymax=0.70,
      ytick={0.45,0.50,...,0.70},
      legend entries={BS,Diverse BS,Top-p,Top-k},
      ylabel style={at={(-.15,0.5)}, scale=0.8},xlabel style={yshift=0.3em,align=center},
      y tick label style={scale=0.8,
        /pgf/number format/fixed,
        /pgf/number format/fixed zerofill,
        /pgf/number format/precision=2},
      bar width=.4cm,
      clip=false,
      ]
      \addplot[fill=red!30, draw=red,area legend] coordinates {(0,0.570) (1,0.558) };
      \addplot[fill=blue!30, draw=blue,area legend] coordinates {(0,0.589) (1,0.577) };
      \addplot[fill=teal!30, draw=teal,area legend] coordinates {(0,0.650) (1,0.631) };
      \addplot[fill=orange!30, draw=orange,area legend] coordinates {(0,0.664) (1,0.645) };
      \node (caption a) at (xticklabel* cs:0.5) {};
    \end{axis}
  \end{scope}
  \node (tmp) at (0,0) {};
  \node [anchor=north] (a) at ([yshift=-0.6cm]caption a|-tmp) {\footnotesize{(a)}};

  \begin{scope}[xshift=\textwidth/3]
    \begin{axis}
      [at={(0,0)},
      ymajorgrids,
      xmajorgrids,
      grid style=dashed,
      width=.33\textwidth,
      height=.28\textwidth,
      symbolic x coords={1 (5K), 2 (10K), 3 (15K), 4 (20K), 5 (25K)},
      xtick=data,
      x tick label style={font=\scriptsize},
      y tick label style={scale=.80,
        /pgf/number format/fixed,
        /pgf/number format/fixed zerofill,
        /pgf/number format/precision=2},
      ymin=0.45, ymax=0.75,
      ytick={0.45,0.50,...,0.75},
      xlabel={Output Size (Sample Size)},
      ylabel={Correlation Score},
      ylabel style={at={(-.15,0.5)},scale=0.8},
      xlabel style={yshift=0.5em, scale=0.8},
      legend entries={\scriptsize{Kend.},  \scriptsize{Kend.}\textsuperscript{+},  \scriptsize{Spear.}, \scriptsize{Spear.}\textsuperscript{+}},
      legend columns=2,
      legend style={
        at={(0.5,1.05)},
        anchor=south,
        legend cell align=left,
        legend plot pos=right,
        /tikz/every even column/.style={column sep=.5cm},
        font=\scriptsize},
      ]
      \addplot [mygreen, mark=triangle*, mark size = 2, line width=0.8pt] coordinates { 
        (1 (5K), 0.498)
        (2 (10K), 0.513)
        (3 (15K), 0.519)
        (4 (20K), 0.521)
        (5 (25K), 0.520)};
      
      \addplot [dashed, mygreen, mark=triangle*, mark size = 2, line width=0.8pt] coordinates { 
        (1 (5K), 0.498)
        (2 (10K), 0.530)
        (3 (15K), 0.544)
        (4 (20K), 0.549)
        (5 (25K), 0.546)};

      \addplot [myred, mark=pentagon*, mark size = 2, line width=0.8pt]coordinates { 
        (1 (5K), 0.628)
        (2 (10K), 0.645)
        (3 (15K), 0.664)
        (4 (20K), 0.670)
        (5 (25K), 0.672)};

      \addplot [dashed, myred, mark=pentagon*, mark size = 2, line width=0.8pt] coordinates { 
        (1 (5K), 0.628)
        (2 (10K), 0.661)
        (3 (15K), 0.676)
        (4 (20K), 0.680)
        (5 (25K), 0.677)};
      \node (caption b) at (xticklabel* cs:0.5) {};
    \end{axis}
    \node [anchor=north] (b) at ([yshift=-.6cm]caption b|-tmp) {\footnotesize{(b)}};
  \end{scope}

  \begin{scope}[xshift=\textwidth*2/3]
    \begin{axis}[
      ymajorgrids,
      grid style=dashed,
      at={(0,0)},
      width=.33\textwidth,
      height=.28\textwidth,
      ybar,
      bar width=0.4cm,
      enlarge x limits=0.20,
      xtick=data,
      xtick align=inside,
      nodes near coords align={vertical},
      ymin=0.20, ymax=0.50,
      ytick={0.20,0.25,...,0.50},
      y tick label style={
        scale=.8,
        /pgf/number format/fixed,
        /pgf/number format/fixed zerofill,
        /pgf/number format/precision=2},
      xticklabels={Kend., Spear., Pear.},
      x tick label style={font=\scriptsize},
      ylabel={Correlation Score},
      ylabel style={at={(-.15,0.5)}, scale=.8},
      xlabel style={yshift=0.5em, font=\scriptsize},
      legend entries={Continuous Score,Star},
      legend style={
        at={(0.5,1.05)},
        anchor=south,
        font=\scriptsize,
        legend cell align=left,
        /tikz/every even column/.append style={column sep=0.10cm}},
      ]
      \addplot [fill=teal!30,draw=teal,area legend] coordinates {(1,0.322) (2,0.420) (3,0.463)};
      \addplot [fill=red!30,draw=red,area legend] coordinates {(1,0.336) (2,0.467) (3,0.481)};
      \node (caption c) at (xticklabel* cs:0.5) {};
    \end{axis}
    \node [anchor=north] (c) at ([yshift=-.6cm]caption c|-tmp) {\footnotesize{(c)}};
  \end{scope}
  
\end{tikzpicture}

%% file: images/rating_vs_ranking.tex
\begin{tikzpicture}
  \begin{scope}
    \begin{axis}[
      ymajorgrids,
      grid style=dashed,
      legend style={at={(0.5,1.05)}, anchor=south, font=\scriptsize,/tikz/every even column/.append style={column sep=.1cm}},
      legend cell align={right},
      legend columns=2,
      ybar,
      enlarge x limits=0.25,
      xtick align=inside,
      height=.22\textwidth,
      width=.26\textwidth,
      ylabel={Correlation Score},
      symbolic x coords={0, 1, 2},
      xtick=data,
      ymin=0.25,
      ymax=0.65,
      nodes near coords align={vertical},
      xticklabels={Kend., Spear., Pear.},
      x tick label style={font=\scriptsize},
      legend entries={Rating, Comparison Ranking},
      legend to name=legendbox,
      ylabel style={at={(0.13,0.5)}, scale=0.8},xlabel style={yshift=0.3em,align=center},
      y tick label style={scale=0.8,
        /pgf/number format/fixed,
        /pgf/number format/fixed zerofill,
        /pgf/number format/precision=2},
      bar width=.3cm,
      clip=false,
      ]
      \addplot[fill=teal!30, draw=teal,area legend] coordinates {(0,0.386) (1,0.529) (2, 0.566)};
      \addplot[fill=red!30, draw=red,area legend] coordinates {(0,0.314) (1,0.435) (2, 0.463)};
      \node (caption a) at (xticklabel* cs:0.5) {};
      \coordinate (p1) at (rel axis cs:1,1);
    \end{axis}
  \end{scope}
  \node (tmp) at (0,0) {};
  \node [anchor=north] (a) at ([yshift=-.4cm]caption a|-tmp) {\footnotesize{(a)}};

  \begin{scope}[xshift=.25\textwidth]
    \begin{axis}[
      ymajorgrids,
      grid style=dashed,
      legend style={at={(0.5,1.05)}, anchor=south, font=\scriptsize,/tikz/every even column/.append style={column sep=.1cm}},
      legend cell align={right},
      legend columns=2,
      ybar,
      enlarge x limits=0.5,
      xtick align=inside,
      height=.22\textwidth,
      width=.26\textwidth,
      ylabel={ChatRef.},
      symbolic x coords={0, 1},
      xtick=data,
      ymin=65.00,
      ymax=75.00,
      nodes near coords align={vertical},
      xticklabels={RL, Reranking},
      x tick label style={font=\scriptsize},
      ylabel style={at={(.08,0.5)}, scale=0.8},xlabel style={yshift=0.3em,align=center},
      y tick label style={scale=0.8,
        /pgf/number format/fixed,
        /pgf/number format/fixed zerofill,
        /pgf/number format/precision=2},
      bar width=.3cm,
      clip=false,
      ]
      \addplot[fill=teal!30, draw=teal] coordinates {(0,73.84) (1,74.01) };
      \addplot[fill=red!30, draw=red] coordinates {(0,71.43) (1,73.54) };
      \node (caption a) at (xticklabel* cs:0.5) {};
      \coordinate (p2) at (rel axis cs:0,1);
    \end{axis}
  \end{scope}
  \node [anchor=south] at ($(p1)!.5!(p2)$) {\pgfplotslegendfromname{legendbox}};
  \node (tmp) at (0,0) {};
  \node [anchor=north] (a) at ([yshift=-.4cm]caption a|-tmp) {\footnotesize{(b)}};

\end{tikzpicture}

%% file: tables/appendix_classification_loss.tex
\begin{table}[t]
    \centering
    \caption{
        Comparisons of our CSEM with reference-based evaluation on different optimization methods. 
    }
    \scalebox{0.84}{
    \begin{tabular}{lccccccc}
    \toprule[1.1pt]
    \multirow{2}{*}{Method}         & \multicolumn{1}{c}{\multirow{2}{*}{\begin{tabular}[c]{@{}c@{}}Evaluation\\ Model\end{tabular}}} & \multicolumn{3}{c}{ChatGPT}                      & \multicolumn{3}{c}{Human}                                \\ \cmidrule(l){3-5} \cmidrule(l){6-8} 
    & \multicolumn{1}{c}{}          & Kend.          & Spear.         & Pear.          & Kend.          & Spear.         & Pear.                   \\ \midrule
    \multirow{2}{*}{Classification} 
    & CSEM-base       & 0.546       &0.649          & 0.635          & 0.307          & 0.478          & 0.483         \\
    & CSEM-large      & 0.575       &0.668          & 0.677          & 0.312          & 0.499          & 0.532         \\ \hdashline
    \multirow{2}{*}{Regression}     
    & CSEM-base       & 0.584     & 0.667      & 0.701       & 0.372          & 0.512          & 0.535   \\
    & CSEM-large      & \textbf{0.613}        & \textbf{0.743}         & \textbf{0.782}          & \textbf{0.386}          & \textbf{0.529}          & \textbf{0.566} \\ \bottomrule[1.1pt]
    \end{tabular}
    }
    \vspace{-1mm}
    \label{tab:classification}
\end{table}

%% file: tables/diff_LLMs.tex
\begin{tabular}{lrcccccc}
\toprule[1.1pt]
    \multirow{2}{*}{Method} & \multirow{2}{*}{\#Params} & \multicolumn{3}{c}{Coherence} & \multicolumn{3}{c}{Relevance} \\ \cmidrule(l){3-5} \cmidrule(l){6-8}
    & &  Kend. & Spear.& Pear. &  Kend. & Spear.& Pear.  \\ \midrule
    ChatGPT   & - &0.407 &0.474 &0.491 &0.378 &0.430 &0.457    \\ 
    GPT-4     & - &0.443 &0.571 &0.602 &0.441 &0.563 &0.584 \\
    LLaMA     & 8B   &0.362  &0.441   &0.457  & 0.334 & 0.403 & 0.426   \\
    Qwen      & 14B  &0.413  &0.486   &0.514  & 0.372 & 0.416  & 0.455   \\ \hdashline
    CSEM (w/ ChatGPT) &355M &0.386 &0.529 & 0.566 & 0.365 & 0.497& 0.531 \\
    CSEM (w/ GPT-4)   &355M &\textbf{0.421} &\textbf{0.584} &\textbf{0.591} &\textbf{0.434} &\textbf{0.540} &\textbf{0.573}  \\
    CSEM (w/ LLaMA)   &355M &0.375  &0.468   &0.489  & 0.353 & 0.462 & 0.483 \\
    CSEM (w/ Qwen)    &355M &0.432  &0.490   &0.534  & 0.388 & 0.468  & 0.492 \\
    \toprule[1.1pt]
    \multicolumn{8}{l}{\parbox{10cm}{The specific versions of the LLMs we use are as follows: GPT-4 (gpt-4-0613), LLaMA (Llama-3.1-8B-Instruct), and Qwen (Qwen2.5-14B-Instruct).}}
\end{tabular}

%% file: tables/general_purpose.tex
\begin{NiceTabular}{lrccc}
\toprule[1.1pt]
Method       &\#Params    & Kend. & Spear. & Pear. \\  \midrule
ChatGPT       & -  &0.424 &0.487  &0.512       \\  \hdashline
Llama-3.1-8B-Instruct &8B  &0.382   &0.423   &0.445       \\
Qwen2.5-14B-Instruct  & 14B    &0.397       &0.439        &0.474       \\  \hdashline
RM-Gemma-2B    &2B  &0.276  &0.332 & 0.351       \\
RM-Gemma-7B    &7B  &0.312  &0.386 & 0.414       \\
Ziya-LLaMA-7B-Reward &7B  &0.325  &0.391 &0.423       \\  \hdashline
CSEM-base       &125M   &0.388  &0.497 &0.522       \\
CSEM-large      &355M   &\textbf{0.414}  &\textbf{0.503} &\textbf{0.537}      \\  
\bottomrule[1.1pt]
\end{NiceTabular}